%% file: paper.tex
\documentclass[journal]{IEEEtran}
\usepackage{cite}
\usepackage{booktabs} %
\usepackage{paralist}
\usepackage{graphicx}
\usepackage{enumitem}
\usepackage[labelformat=simple]{subcaption}

\usepackage[usenames,dvipsnames,svgnames,table]{xcolor}
\usepackage{amsmath,amssymb,amsfonts}
\usepackage[vlined,ruled,linesnumbered]{algorithm2e}
\usepackage{upgreek}
\usepackage{makecell}
\setlist[itemize]{nosep,leftmargin=1em}  %
\setlist[enumerate]{nosep,leftmargin=1.2em}  %
\usepackage{xspace}
\usepackage{multirow}
\usepackage{listings}
\usepackage{kotex}
\usepackage{pdfrender}
\usepackage[most]{tcolorbox}
\usepackage{footmisc}
\usepackage{diagbox}
\usepackage{hyperref}
\usepackage{cleveref}
\hypersetup{
	pdftitle={Knowledge-Guided Dynamic Systems Modeling: A Case Study on Modeling River Water Quality},
	pdfauthor={Namyong Park, MinHyeok Kim, Nguyen Xuan Hoai, R.I. (Bob) McKay, and Dong-Kyun Kim},
	pdfkeywords={dynamic system modeling, model revision, prior knowledge incorporation, river water quality modeling, evolutionary algorithm}
}

\input{dfn}

\begin{document}

\title{Knowledge-Guided Dynamic Systems Modeling:\\ A Case Study on Modeling River Water Quality}

\author{Namyong~Park,
MinHyeok~Kim,
Nguyen~Xuan~Hoai,
R.I.~(Bob)~McKay,
and~Dong-Kyun~Kim%
\thanks{Namyong Park is with the Computer Science Department, Carnegie Mellon University, Pittsburgh, PA, USA (e-mail: namyongp@cs.cmu.edu).}%
\thanks{MinHyeok Kim is with LG Electronics, Seoul, Korea (e-mail: min.fourleaf@gmail.com).}%
\thanks{Nguyen Xuan Hoai is with AI Academy Vietnam, Hanoi, Vietnam (e-mail: nxhoai@aiacademy.edu.vn).}%
\thanks{R.I. (Bob) McKay is with the College of Engineering \& Computer Science, Australian National University, Canberra, Australia (e-mail: rimanucs@gmail.com).}%
\thanks{Dong-Kyun Kim is with K-water Research Institute, Daejeon, Korea (e-mail: dkkim1004@gmail.com).}%
}

\markboth{}%
{Park \MakeLowercase{\textit{et al.}}: Knowledge-Guided Dynamic Systems Modeling: A Case Study on Modeling River Water Quality}

\maketitle

\begin{abstract}
Modeling real-world phenomena is a focus of many science and engineering efforts,
such as ecological modeling and financial forecasting, to name a few.
Building an accurate model for complex and dynamic systems improves understanding of 
underlying processes and leads to resource efficiency.
Towards this goal, knowledge-driven modeling builds a model based on human expertise, yet is often suboptimal.
At the opposite extreme, data-driven modeling learns a model directly from data, requiring extensive data and potentially generating overfitting.\hide{, and models inconsistent with prior knowledge.}
We focus on an intermediate approach, model revision, in which prior knowledge and data are combined
to achieve the best of both worlds.
In this paper, we propose a genetic model revision framework 
based on tree-adjoining grammar (TAG) guided genetic programming (GP),
using the TAG formalism and GP operators in an effective mechanism
to incorporate prior knowledge and make data-driven revisions in a way that complies with prior knowledge.
Our framework is designed to address the high computational cost of evolutionary modeling of complex systems.
Via a case study~on the challenging problem of river water quality modeling,
we show that the framework efficiently learns an interpretable model, with higher modeling accuracy than existing methods.
\end{abstract}


\IEEEpeerreviewmaketitle

\vspace{-1em}
\section{Introduction}
\label{sec:intro}
\input{010intro}

\section{River Water Quality Modeling}
\label{sec:prelim}
\input{020background}

\vspace{-0.3em}
\section{Methods}
\vspace{-0.3em}
\label{sec:methods}
\input{030methods}

\section{Experiments}
\label{sec:exp}
\input{040experiments}
\section{Related Work}
\label{sec:relatedwork}

\input{050relatedwork}
\section{Conclusion}
\label{sec:conclusion}
\input{060conclusion}

\appendices
\crefalias{section}{appendix}

\input{070appendix}

\vspace{-0.5em}
\bibliographystyle{IEEEtran}

\end{document}

%% file: dfn.tex
\definecolor{Gray}{gray}{0.9}
\definecolor{LightGray}{gray}{0.96}
\definecolor{LightCyan}{rgb}{0.88,1,1}
\newcommand{\best}[1]{\textbf{\cellcolor{Gray}#1}}

\newcommand\dunderline[3][-1pt]{{%
		\sbox0{#3}%
		\ooalign{\copy0\cr\rule[\dimexpr#1-#2\relax]{\wd0}{#2}}}}

\newcommand*{\belowrulesepcolor}[1]{%
	\noalign{%
		\kern-\belowrulesep
		\begingroup
		\color{#1}%
		\hrule height\belowrulesep
		\endgroup
	}%
}
\newcommand*{\aboverulesepcolor}[1]{%
	\noalign{%
		\begingroup
		\color{#1}%
		\hrule height\aboverulesep
		\endgroup
		\kern-\aboverulesep
	}%
}

\newcommand*{\boldcheckmark}{%
	\textpdfrender{
		TextRenderingMode=FillStroke,
		LineWidth=.7pt, %
	}{\checkmark}%
}

\newcolumntype{g}{>{\columncolor{Gray}}c}

\newcommand{\hide}[1]{}

\newcommand{\highlight}[1]{\textcolor{blue}{#1}}

\newcommand{\eat}[1]{}

\newcommand{\bci}{\begin{compactitem}}
	\newcommand{\eci}{\end{compactitem}}
\newcommand{\bce}{\begin{compactenum}}
	\newcommand{\ece}{\end{compactenum}}

\newcommand{\method}{\textsc{GMR}\xspace}
\newcommand{\gggp}{\textsc{GGGP}\xspace}

\newcommand{\ga}{\textsc{GA}\xspace}
\newcommand{\mc}{\textsc{MC}\xspace}
\newcommand{\lhs}{\textsc{LHS}\xspace}
\newcommand{\mle}{\textsc{MLE}\xspace}
\newcommand{\mcmc}{\textsc{MCMC}\xspace}
\newcommand{\sa}{\textsc{SA}\xspace}
\newcommand{\dream}{\textsc{DREAM}\xspace}
\newcommand{\sceua}{\textsc{SCE-UA}\xspace}
\newcommand{\demcz}{\textsc{DE-MC\textsubscript{Z}}\xspace}

\newcommand{\rnn}{\textsc{RNN}\xspace}
\newcommand{\rnnbusan}{\textsc{RNN-S1}\xspace}
\newcommand{\rnnall}{\textsc{RNN-All}\xspace}

\newcommand{\arimax}{\textsc{ARIMAX}\xspace}
\newcommand{\arimaxBusan}{\textsc{ARIMAX-S1}\xspace}
\newcommand{\arimaxAll}{\textsc{ARIMAX-All}\xspace}

\newcommand{\manual}{\textsc{Manual}\xspace}

%% file: 010intro.tex
\IEEEPARstart{M}{odeling} real-world phenomena is the goal of 
numerous science and engineering endeavors,
such as ecological modeling~\cite{DBLP:conf/cec/KimMSLN10}, financial forecasting~\cite{lu2009financial}, 
user modeling~\cite{DBLP:journals/umuai/WebbPB01}, disease prediction~\cite{10.1371/journal.pone.0199839},
popularity estimation~\cite{DBLP:conf/kdd/ParkKDZF19,DBLP:conf/kdd/ParkKDZF20},
student dropout prediction~\cite{DBLP:journals/corr/abs-2002-01598},
and drug discovery~\cite{becker2003modeling}.
An accurate model of these systems can enable better understanding of underlying mechanisms
and more effective use of resources.
Real-world systems are typically dynamic and complex, 
with multiple observed and latent variables 
that change over time, and affect each other in complex and often nonlinear ways.
As an example, consider the task of forecasting river water quality.
Addressing this problem requires an understanding of processes such as plankton dynamics and 
hydrological mechanisms, and modeling how they influence the system dynamics as a whole.

Existing approaches for modeling dynamical systems can be grouped into three classes.
The first is \textit{knowledge-driven modeling}.
The structure of knowledge-driven models and their parameters are determined by domain experts, 
based on their prior knowledge and using observational data to calibrate the model parameters.
In knowledge-driven modeling, the state of dynamic systems can be modeled by differential equations.
While knowledge-driven models perform reasonably well when the modelled system is simple, they take time to construct,
and generally perform less well with increasing system complexity.

The second is \textit{data-driven modeling}:
learning a model purely from data, with no need for prior knowledge.
Highly accurate models can often be obtained by these methods.
Modeling complex systems requires plentiful data, 
but the high cost of measurement~\cite{DBLP:journals/tkde/KarpatneAFSBGSS17} means this is often unavailable.
Sadly, learning a model from limited data often leads to overfitting.
Importantly, data-driven modeling might learn models that are not consistent with prior knowledge.
Also, some of the popular methods in this class (e.g., neural networks)  generate black box models, lacking explanatory power.

The third class combines knowledge- and data-driven modeling to gain the best of both worlds.
\textit{Model calibration} is one widely used approach: the initial model structure is specified by domain knowledge, and then model parameters are optimized using data.
However, model calibration updates only the model parameters, not the model structure. 
If this is oversimplified, the accuracy of the optimized model will be compromised, and the calibrated parameter values will be unrealistic.
\textit{Model revision} is a more interesting and effective approach: 
prior knowledge specifies the initial model structure and parameter values,
but both are updated iteratively to obtain a better fit to the data.
This approach of revising and improving existing models 
closely resembles traditional scientific discovery process~\cite{dvzeroski2007computational}.
\textit{Knowledge-guided model revision} further improves plain model revision
by letting model revision be guided by prior knowledge 
and producing a revised model consistent with domain knowledge.
\Cref{tab:modeling_approaches} summarizes how different approaches satisfy 
desirable properties for knowledge-guided modeling of complex dynamic systems.

\begin{table}[!t]
\par\vspace{-1.0em}\par
\setlength{\tabcolsep}{0.3mm}
\renewcommand{\aboverulesep}{1.0pt}
\renewcommand{\belowrulesep}{1.0pt}
\centering
\caption{Model revision satisfies all properties for interpretable knowledge-guided modeling of complex dynamic systems.
	Other approaches miss one or more of the properties. ``?'' means that it depends on the specific method used.}
\makebox[0.4\textwidth][c]{\setlength\doublerulesep{0.5pt}
	\begin{tabular}{l | c c c c g}
		\toprule
		\diagbox{\textbf{Property}}{\textbf{Approach}} & \textbf{\rotatebox[origin=c]{0}{\makecell{Knowledge-\\Driven\\Modeling}}} & \textbf{\makecell{\rotatebox[origin=c]{0}{\makecell{Data-\\Driven\\Modeling}}}} & \textbf{\rotatebox[origin=c]{0}{\makecell{Model\\Calibration}}} & \textbf{\rotatebox[origin=c]{0}{\makecell{Model\\Revision}}} & \textbf{\rotatebox[origin=c]{0}{\makecell{Knowledge-\\Guided\\Model\\Revision}}} \\ \midrule 
		\makecell[l]{Learning models\\ consistent with\\ prior knowledge} & $\checkmark$ & & ? & & $\boldcheckmark$ \\ \midrule
		\makecell[l]{Knowledge-based\\model specification} & $\checkmark$ & & $\checkmark$ & $\checkmark$ & $\boldcheckmark$ \\ \midrule
		\makecell[l]{Structural\\ model update} & & ? & & $\checkmark$ & $\boldcheckmark$ \\ \midrule
		\makecell[l]{Automatic tuning of\\ model parameters} & & $\checkmark$ & $\checkmark$ & $\checkmark$ & $\boldcheckmark$ \\ \midrule
		\makecell[l]{Capacity to model\\complex systems} & & $\checkmark$ & & $\checkmark$ & $\boldcheckmark$ \\ \midrule
		\makecell[l]{Interpretable} & $\checkmark$ & ? & $\checkmark$ & $\checkmark$ & \best{$\boldcheckmark$} \\ 
		\bottomrule
	\end{tabular}
}
\label{tab:modeling_approaches}
\vspace{-2em}
\end{table}

As a powerful technique for evolving programs,
genetic programming (GP)~\cite{DBLP:books/daglib/0070933} provides an effective framework for model revision.
GP has been successfully applied to real-world problems in various fields~\cite{DBLP:reference/genetic/VyasGT15,DBLP:books/daglib/0023033,DBLP:conf/gecco/PhongHMSUP12}, 
and has the theoretical advantage that the output is interpretable, unlike blackbox models.
Among GP's methods, symbolic regression (SR), which aims to discover a function that fits the training data, is the most relevant to process modeling.
Standard SR is a form of data-driven modeling, as it sets no restrictions on the model structure. It thus suffers from a lack of guidance in the optimization process, and may produce models that violate domain knowledge.

A number of newer GP methodologies, such as grammar guided GP (GGGP)~\cite{Whigham:1996:GBE:923829} and tree-adjoining grammar (TAG) guided GP (TAG3P)~\cite{hoaithesis2004flexible}, support constraining or biasing the structure of learnt models~\cite{DBLP:journals/ec/Montana95,o2001grammatical}.
We base our framework on TAG3P,
which is a powerful tool for incorporating domain knowledge 
while exploring the complex 
search spaces required for modeling real-world processes.

In this paper, we propose TAG3P-based genetic model revision (\method), 
in which the TAG formalism and GP operators provide an effective mechanism 
to perform data-driven model revisions based on prior knowledge.
We show how to represent dynamic processes in TAG,
and how to extend the TAG3P framework to incorporate different types of prior knowledge into the optimization process.
An important challenge in applying GP to complex systems is
the high computational cost of the search and fitness evaluation in GP systems.
Our framework achieves efficient and effective optimization
by reducing redundancy and enabling evaluation short-circuiting.
In our case study, \method allows us to accurately model water quality in a river ecosystem,
a complex dynamic system with extensive geographic coverage, which has previously been much less studied than relatively simple lake ecosystems 
due to its far higher complexity.

In summary, our contributions are as follows:
\begin{itemize}[leftmargin=1em]
\item \textit{Framework.} We present a \method framework for dynamic systems modeling,
which improves a knowledge-based model in a data-driven manner, guided by prior knowledge.
\item \textit{Knowledge Incorporation.} We design novel mechanisms to represent prior knowledge and
perform knowledge-guided optimizations in the \method framework. %
\item \textit{River Modeling.}
This is the first work to apply model revision to modeling a river system.
Previous work on river modeling used model calibration alone.
\item \textit{Effectiveness.} Our framework achieves 
the best forecasting accuracy in river modeling among a variety of methods,
while producing models consistent with domain knowledge.
\item \textit{Efficiency.} We present techniques to cut down the computational cost of GP systems, achieving $607\times$ speedup.
\end{itemize}
\textbf{Reproducibility}: Code and data are available at \url{https://www.cs.cmu.edu/~namyongp/gmr}.

\begin{figure}
	\par\vspace{-0.5em}\par
	\centering
	\makebox[0.5\linewidth][c]{\includegraphics[width=1.06\linewidth]{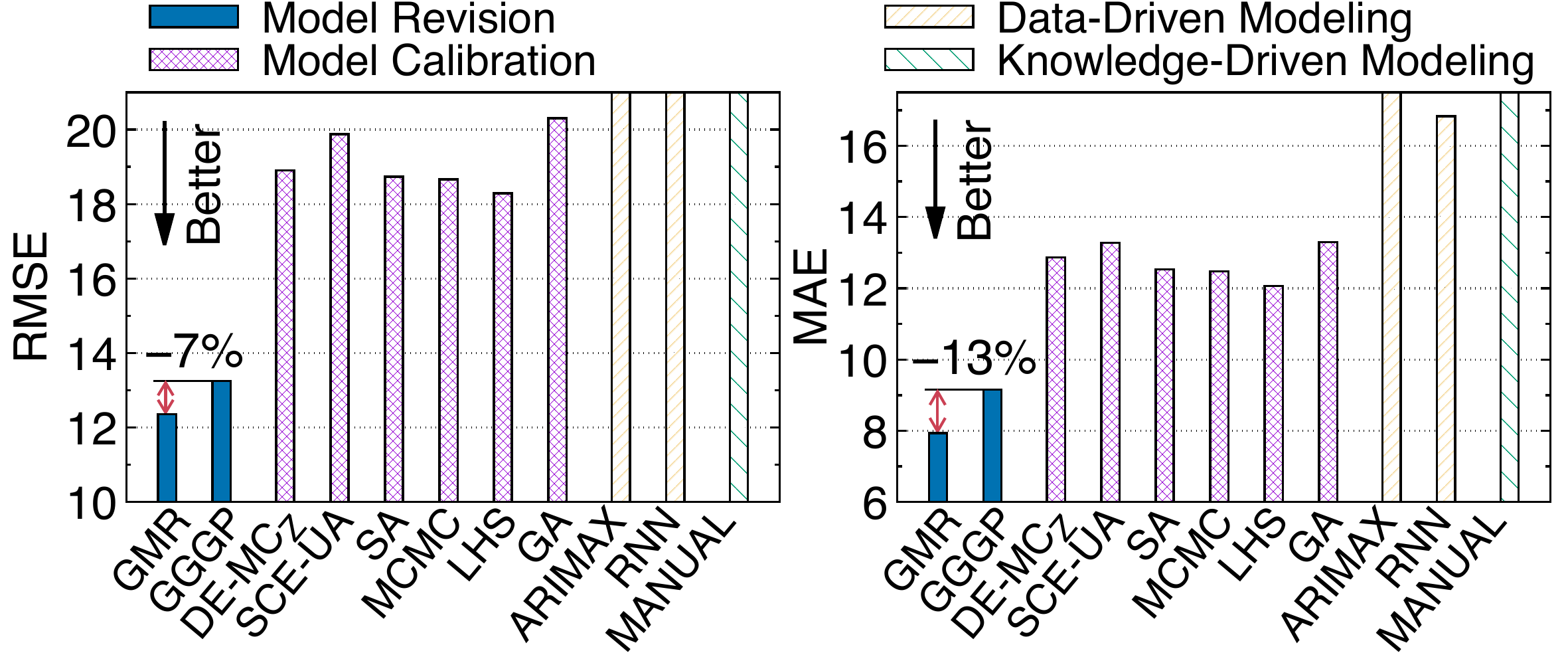}}
	\caption{\method achieves the best forecasting accuracy in the river modeling task,
		obtaining 7\% and 13\% lower RMSE (left) and MAE (right), respectively, than the second best method, 
		while producing revised models guided by domain knowledge. 
	}
	\label{fig:crownjewel}
	\vspace{-1em}
\end{figure}

The rest of the paper is organized as follows.
We describe the river modeling problem in~\Cref{sec:prelim}, and 
present the \method framework and how to apply it to river modeling in~\Cref{sec:methods}.
After presenting experimental results in~\Cref{sec:exp}, we review related works in~\Cref{sec:relatedwork}, and conclude in~\Cref{sec:conclusion}.

%% file: 020background.tex
Rivers are precious freshwater resources for households, farming, and industry.
Due to intensive use and increasing development,
the eutrophication (over-enrichment with nutrients) of rivers  
has become a serious global problem.
Algal blooms are one of the most problematic and widespread consequences that deteriorate river water quality~\cite{moss2009ecology}.
For improved river management, it is crucial to have an accurate model of the water quality.

River water quality modeling aims to predict phytoplankton biomass,
a proxy for eutrophication.
Based on the knowledge of a freshwater ecologist, 
we designed the following biological processes, modeling the change of phytoplankton biomass over time
by capturing the interplay between phytoplankton ($B_{Phy}$) and zooplankton ($ B_{Zoo} $).

{%
\footnotesize
\setlength{\abovedisplayskip}{1pt}
\setlength{\belowdisplayskip}{\abovedisplayskip}
\setlength{\abovedisplayshortskip}{0pt}
\setlength{\belowdisplayshortskip}{1pt}
\begin{align}
\label{eq:biologicalP:raw}
\frac{d {B_{\mathit{Phy}}}}{d t}	& = B_\mathit{Phy} \cdot ({\mu}_\mathit{Phy} - {\gamma}_\mathit{Phy}) - B_\mathit{Zoo} \cdot \varphi\\
\mu_{\mathit{Phy}}	& = C_\mathit{UA} \cdot f(V_\mathit{lgt}) \cdot g(V_{n},V_{p},V_{si}) \cdot h(V_\mathit{tmp}) \nonumber \\
\gamma_{\mathit{Phy}}	& = C_{\mathit{BRA}} \nonumber\\
\varphi		& = C_\mathit{MFR} \cdot \mathit{\lambda_{Phy}} \nonumber\\ 
\mathit{\lambda_{Phy}}	& = ({B_\mathit{Phy}}-C_\mathit{Fmin}) / (C_\mathit{FS}+B_\mathit{Phy} - C_\mathit{Fmin}) \nonumber\\
f(V_{\mathit{lgt}})	& = (V_{\mathit{lgt}} / C_{\mathit{BL}}) \cdot e^{1- (V_{\mathit{lgt}} / C_{\mathit{BL}})}    \nonumber\\
g(V_{n},V_{p},V_{\mathit{si}})	& =\min\left( V_{n} / (C_{N}+V_{n}), V_{p} / (C_{P}+V_{p}), V_{si} / (C_{SI}+V_{si}) \right) \nonumber\\
h(\mathit{V_{tmp}})	& = \max( e^{-C_{\mathit{PT}} (V_{\mathit{tmp}} - C_{\mathit{BTP1}})^2}, e^{-C_{\mathit{PT}} (V_{\mathit{tmp}} - C_{\mathit{BTP2}})^2} )\nonumber\\[0.5em]
\small
\label{eq:biologicalZ:raw}
\frac{d {B_{\mathit{Zoo}}}}{d t}	& = {B_{\mathit{Zoo}}} \cdot ({\mu}_{\mathit{Zoo}} - {\gamma}_\mathit{Zoo} - {\delta}_\mathit{Zoo} ) \\ 
{\mu}_\mathit{Zoo}	& = C_\mathit{UZ} \cdot \mathit{\lambda_{Phy}} \nonumber \\
{\gamma}_\mathit{Zoo}	& = C_\mathit{BRZ} + C_\mathit{BMT} \cdot \varphi \nonumber	\\
{\delta}_\mathit{Zoo}	& = C_\mathit{DZ} \nonumber
\end{align}}%

\begin{figure*}[h!]
	\centering
	\begin{subfigure}[b]{0.45\textwidth}
		\includegraphics[trim={0 0.40cm 0 0.55cm},clip,width=\textwidth]{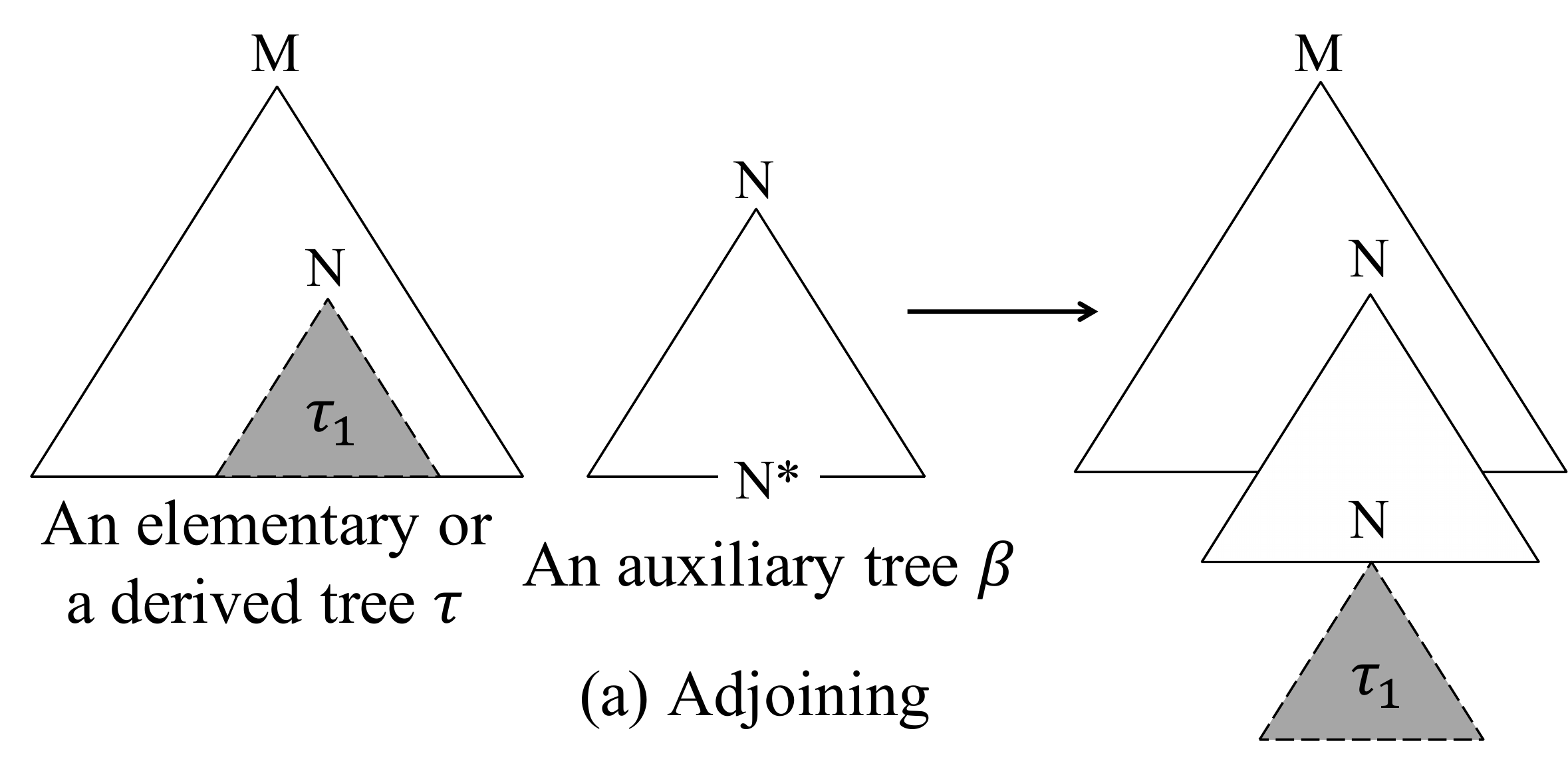}
		\label{fig:adjoining}
	\end{subfigure}
	~~ %
	\begin{subfigure}[b]{0.45\textwidth}
		\includegraphics[trim={0 0.2cm 0 0},clip,width=\textwidth]{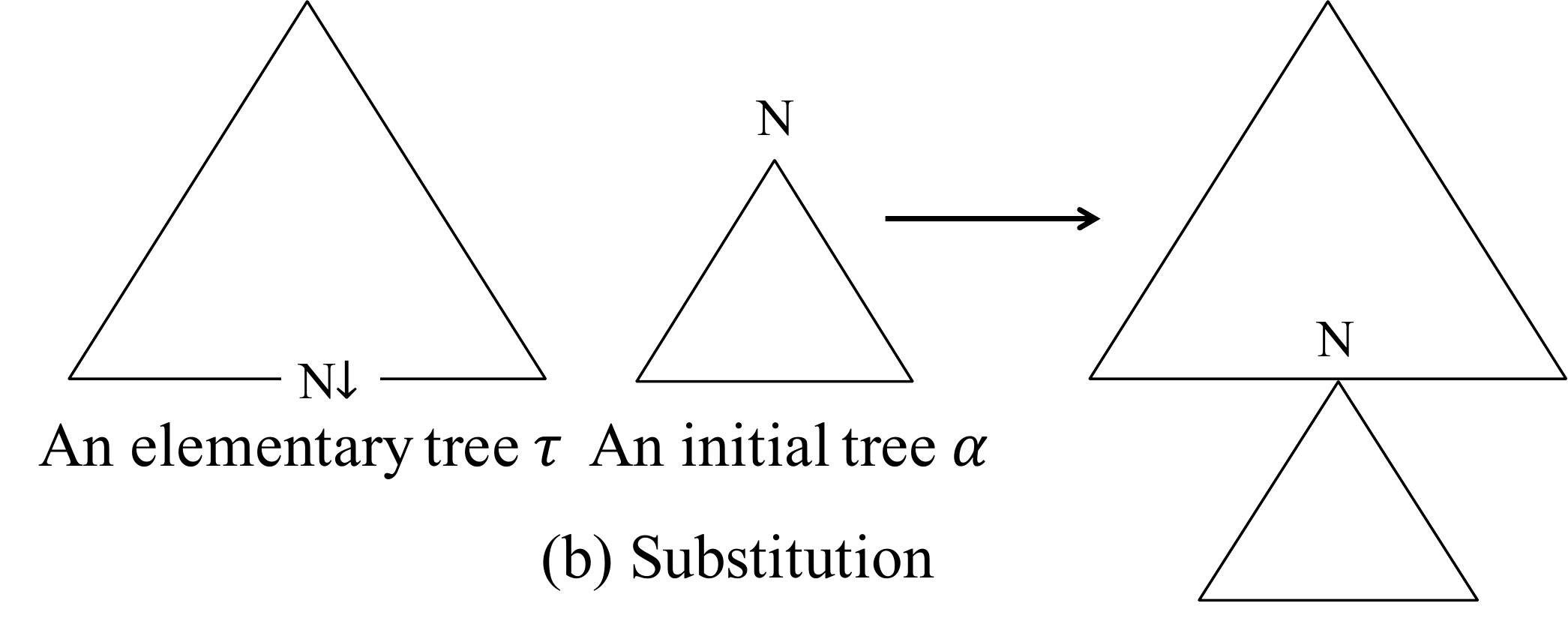}
		\label{fig:substitution}
	\end{subfigure}
	~ %
	\setlength{\abovecaptionskip}{-9pt}
	\caption{Illustrations of tree composition operations used by tree-adjoining grammar (TAG): (a) adjoining and (b) substitution.}
	\label{fig:adjoining_substitution}
	\vspace{-1em}
\end{figure*}

The phytoplankton dynamics model ($ {d {B_{\mathit{Phy}}}}/{d t} $) incorporates 
the photosynthetic productivity ($ {\mu}_\mathit{Phy} $), metabolic degradation ($ {\gamma}_\mathit{Phy} $), and grazing pressure of zooplankton ($ \varphi $).
The photosynthetic productivity depends on multiplicative influences from  variables such as
light intensity ($ V_{lgt} $), nutrient (nitrogen, phosphorus, and silica) concentrations ($ V_{n}, V_{p}, V_{si} $), and water temperature ($ V_{tmp} $).
These functions build on earlier studies on modeling algal dynamics including \cite{cho1998growth,hongping2002study}. 
Further, considering the effect of summer cyanobacteria and winter diatom blooms,
we extend the process with two additional parameters reflecting optimal temperatures ($ C_{BTP1}, C_{BTP2} $).
The zooplankton dynamics model ($ {d {B_{\mathit{Zoo}}}}/{d t} $),
adapted from \cite{hongping2002study},
incorporates
the growth ($ {\mu}_{Zoo} $), respiration ($ {\gamma}_{Zoo} $), and death ($ {\delta}_{Zoo} $) rates of zooplankton.

The parameters of these biological processes fall into two classes: 
\textit{constant parameters} (starting with $ C $) have constant values representing physiological rates 
(e.g., growth or feeding rate), 
while \textit{variable parameters} (starting with $ V $) correspond to external conditions and forces, changing over time.
In evaluating \eqref{eq:biologicalP:raw} and \eqref{eq:biologicalZ:raw}, 
variable parameters are imported from the observed data at the evaluation time $ t $.
More details on these constant and variable parameters are given in~\Cref{tab:process_variables,tab:process_constants}.

The goal of model revision for our task is summarised as:
\begin{tcolorbox}[boxsep=1pt,boxrule=0pt,left=0.05cm,right=0.05cm,top=0.08cm,bottom=0.08cm]
Given biological processes \eqref{eq:biologicalP:raw} and \eqref{eq:biologicalZ:raw}, 
make relevant changes to the structure and constant parameter values of \eqref{eq:biologicalP:raw} and \eqref{eq:biologicalZ:raw}
guided by prior knowledge such that the estimated phytoplankton biomass ($ B_{\mathit{Phy}} $) is close to the observed values, and 
the revised process is consistent with prior knowledge.
\end{tcolorbox}

River modeling is a challenging task.
Although carefully built with domain knowledge, manually-designed processes \eqref{eq:biologicalP:raw} and \eqref{eq:biologicalZ:raw} (\manual) exhibit poor predictive performance, as shown in~\Cref{fig:crownjewel}.
While the results of model calibration methods show that parameter tuning greatly improves modeling accuracy,
it is not enough to be able to update only the model parameters.
By improving the process itself via model revision,
our method obtains the best result, 
with 13\% and 34\% smaller MAE than is obtained with the second best method and the best model calibration result, respectively,
while producing revised processes guided by prior knowledge.

%% file: 030methods.tex
In this section, we present our genetic model revision (\method) framework.
There are three major challenges in applying model revision to the modeling of dynamical systems.
\begin{enumerate}[label=\arabic*)]
	\item \textbf{Representation of dynamic processes.}
	Given differential equations that model dynamic processes, such as the one underlying river water quality (\eqref{eq:biologicalP:raw} and \eqref{eq:biologicalZ:raw}),
	how can we represent them for successful model revision?
	\item \textbf{Mechanism for knowledge-guided model revision.}
	Model revision requires defining specific steps for making revisions to obtain a better model.
	How can we effectively perform model revision guided by prior knowledge?
	\item \textbf{Efficient and effective model revision.}
	Real-world systems are complex, often incurring high computational cost.
	How can we perform efficient and effective model revision?
\end{enumerate}
In the \method framework, we address these challenges with the following ideas.

\begin{enumerate}[label=\arabic*)]
	\item \textbf{Using tree-adjoining grammar (TAG) for representing dynamic processes}
	provides a powerful framework to succinctly express dynamic processes and their revision,
	while facilitating controlled incorporation of prior knowledge.
	\item \textbf{Making revision via TAG-guided GP and expressing prior knowledge using the TAG formalism}
	leads to an accurate model consistent with prior knowledge.
	\item \textbf{Removing redundancy, speeding up operations, and local search} enable fast and effective model revision.

\end{enumerate}
We describe how to represent dynamic processes using TAG in \Cref{sec:methods:tag}, and 
present the \method framework in \Cref{sec:methods:framework}.
Then we demonstrate how to apply \method to real-world problems, such as river modeling, in \Cref{sec:methods:rivermodeling},
and present techniques to improve efficiency and effectiveness in \Cref{sec:methods:improving}.

\vspace{-1em}
\subsection{Representing Dynamic Processes Using TAG}
\label{sec:methods:tag}

\subsubsection{Preliminaries on TAG (Tree-Adjoining Grammar)}
\begin{figure}[t!]
	\par\vspace{-0.5em}\par
	\centering
	\makebox[0.5\linewidth][c]{\includegraphics[trim={0 0 0 0},clip,width=1.1\linewidth]{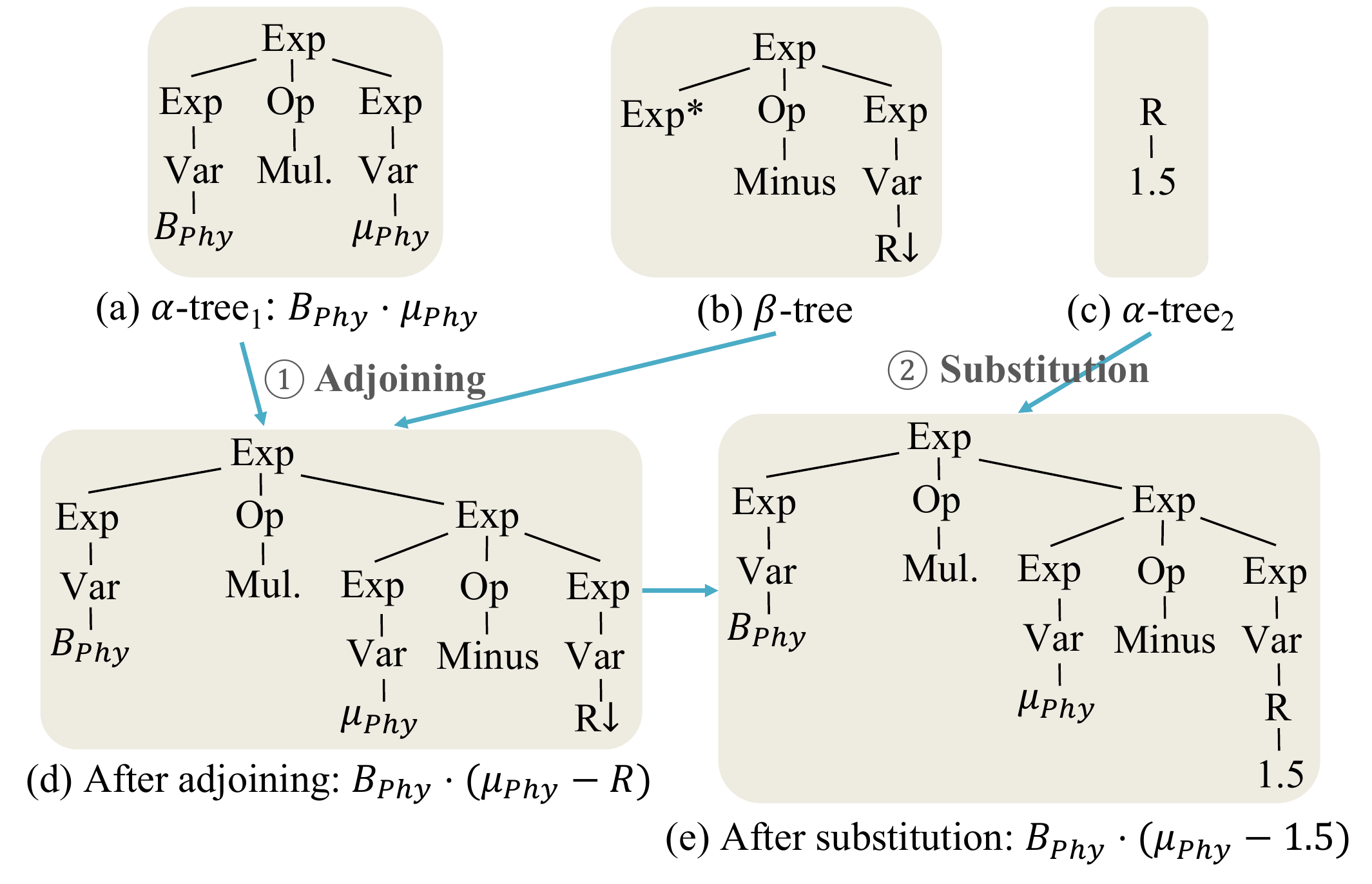}}
	\setlength{\abovecaptionskip}{0pt}
	\caption{(a)--(c): Example $ \alpha $- and $ \beta $- trees representing a dynamic process and potential revisions.
		(d), (e): Resulting trees after adjoining and substitution (see text for details).}
	\label{fig:tag_example}
	\vspace{-1.0em}
\end{figure}

A TAG is a tree generating system~\cite{joshi1997tree},
consisting of a quintuple ($T$, $N$, $I$, $A$, $S$) where
\begin{itemize}
	\item $T$ is a finite set of terminal symbols;
	\item $N$ is a finite set of non-terminal symbols ($N \cap T = \emptyset$);
	\item $S \in N$ is a non-terminal symbol called the \textit{start symbol};
	\item $I$ is a set of finite trees called \textit{initial trees} or \textit{$\alpha$-trees};
	\item $A$ is a set of finite trees called \textit{auxiliary trees} or \textit{$\beta$-trees}.
\end{itemize}
\Cref{fig:tag_example,fig:adjoining_substitution} provide illustrations of $ \alpha $- and $ \beta $-trees.
The trees in $I \cup A$ (i.e., $ \alpha $- and $ \beta $-trees) are referred to as \textit{elementary trees}.
In an elementary tree, the labels of all interior nodes are non-terminal symbols, 
while the labels of the nodes on the frontier can be either terminal or non-terminal symbols.
The frontier nodes of an elementary tree with non-terminal symbols are marked as $\downarrow$ for substitution,
except for one special node in an auxiliary tree, which is called the \textit{foot node} and marked with an asterisk ($*$) by convention.
The foot node must have the same non-terminal symbol as that of the corresponding tree's root node (e.g., see \Cref{fig:tag_example}(b)).

\textit{Adjoining} and \textit{substitution} are the two composition operations TAG uses to construct a \textit{derived} tree (\Cref{fig:adjoining_substitution}).
Adjoining builds a new tree given an auxiliary tree $\beta$ and a tree $\tau$ (which can be either an elementary or a derived tree). 
Assume that the root of $\beta$ is labeled as $N$, and that the tree $\tau$ has an interior node $n$ labeled as $N$.
The steps for adjoining $\beta$ into $\tau$ are as follows (see \Cref{fig:adjoining_substitution}(a) for an illustration):
\begin{enumerate}[label=\arabic*)]
	\item The sub-tree $\tau_1$ rooted at node $n$ is disconnected from $\tau$;
	\item The tree $\beta$ is attached at the place where the node $n$ was;
	\item $\tau_1$ is attached to the foot node (marked with $*$) of the tree~$\beta$.
\end{enumerate}

Substitution creates a derived tree from an elementary tree $\tau$ and 
a tree $\alpha$ which is (derived from) an initial tree.
As in \Cref{fig:adjoining_substitution}(b),
substitution selects a non-terminal on the frontier of the tree $\tau$ (marked as $\downarrow$)
matching the root of $\alpha$, 
and replaces it with~$\alpha$.
A tree derived from an initial tree, and lacking frontier non-terminals, is a \textit{completed} tree.

An in-depth description of TAG appears in~\cite{joshi1997tree,hoaithesis2004flexible}.

\begin{figure}[t!]
	\centering
	\includegraphics[width=1.7in]{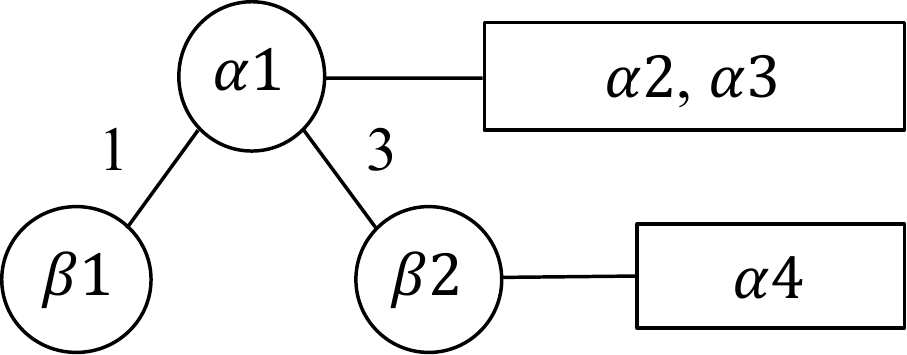}
	\caption{TAG derivation tree which encodes a revised differential equation. Two nodes (labeled by $\beta1$ and $\beta2$) are $\beta$-trees 
		that are adjoined into the specified address (the number on the link) of the root. 
		Rectangles contain $\alpha$-trees (lexemes), which are substituted into the open nodes in the linked tree.
	}
	\label{fig:dtree}
	\vspace{-1.0em}
\end{figure}

\subsubsection{TAG-Based Dynamic Process Representation}
Consider this equation, a simplified form of \eqref{eq:biologicalP:raw}, to see how TAG can represent dynamic processes and potential revisions:
\begin{align}\label{eq:simple_process}
\frac{d {B_{\mathit{Phy}}}}{d t}	& = B_\mathit{Phy} \cdot {\mu}_\mathit{Phy}
\end{align}
\eqref{eq:simple_process} can be represented by an $ \alpha $-tree shown in~\Cref{fig:tag_example}(a) where ``Mul.'' denotes multiplication.
\Cref{fig:tag_example}(b) shows a $ \beta $-tree representing one potential extension 
where an expression (denoted by ``Exp'') is extended by deducting a random variable (denoted by ``R'') from it.
Then adjoining the $ \beta $-tree in \Cref{fig:tag_example}(b) into the rightmost ``Exp'' node of the $ \alpha $-tree in \Cref{fig:tag_example}(a)
yields the tree shown in \Cref{fig:tag_example}(d), which corresponds to $ B_\mathit{Phy} \cdot ({\mu}_\mathit{Phy} - R) $.
Another $ \alpha $-tree in \Cref{fig:tag_example}(c) encodes a potential value for variable R.
By substituting it into the frontier node R (marked with $ \downarrow $) shown in \Cref{fig:tag_example}(d), 
we obtain a revised process: 
\begin{align}
\frac{d {B_{\mathit{Phy}}}}{d t}	& = B_\mathit{Phy} \cdot ({\mu}_\mathit{Phy} - 1.5).
\end{align}

A completed tree (e.g., \Cref{fig:tag_example}(e))
corresponds to a revised process.
The history of adjunctions and substitutions is encoded as an object tree called the \textit{derivation tree}.
In other words, we encode successive model revisions and the revised process as a derivation tree in TAG.
Among several proposed definitions of TAG derivation tree, 
we use the formulation with restricted substitution~\cite{hoaithesis2004flexible}:
\begin{enumerate}[label=\arabic*)]
\item The root node is labeled with an $\alpha$-tree (i.e., input process) whose root node is labeled by the start symbol $S$.
\item All other nodes are labeled with $\beta$-trees (adjunction nodes). 
An adjunction node is associated with an address of the node at which the adjunction took place.
\item An $ \alpha $-tree that is substituted is restricted to have no children, 
which allows us to regard substitution as an in-node operation, and also simplifies the derivation tree greatly.
\end{enumerate}
With this definition, the TAG derivation tree in \method (\Cref{fig:dtree}) is a tree of objects
where links between objects indicate adjunction at the
specified address, and 
each node has a list of $ \alpha $-trees (called \textit{lexemes}) to be substituted into the open nodes (called \textit{lexicons})
in the elementary tree labeled by the node.

Note that while the above discussion describes how TAG provides a mechanism for representing and revising dynamic processes,
we need a more careful design of $ \alpha $- and $ \beta $-trees than is shown in~\Cref{fig:tag_example},
to be able to make controlled changes.
For example, in order to reflect prior knowledge, we may want to adjoin the $ \beta $-tree in \Cref{fig:tag_example}(b) into only one of the Exp nodes in \Cref{fig:tag_example}(a).
However, with the $ \beta $-tree in \Cref{fig:tag_example}(b), adjoining can happen at any of them.

\begin{figure}[t!]
	\centering
	\includegraphics[trim={0 0 0 0},clip,width=0.95\linewidth]{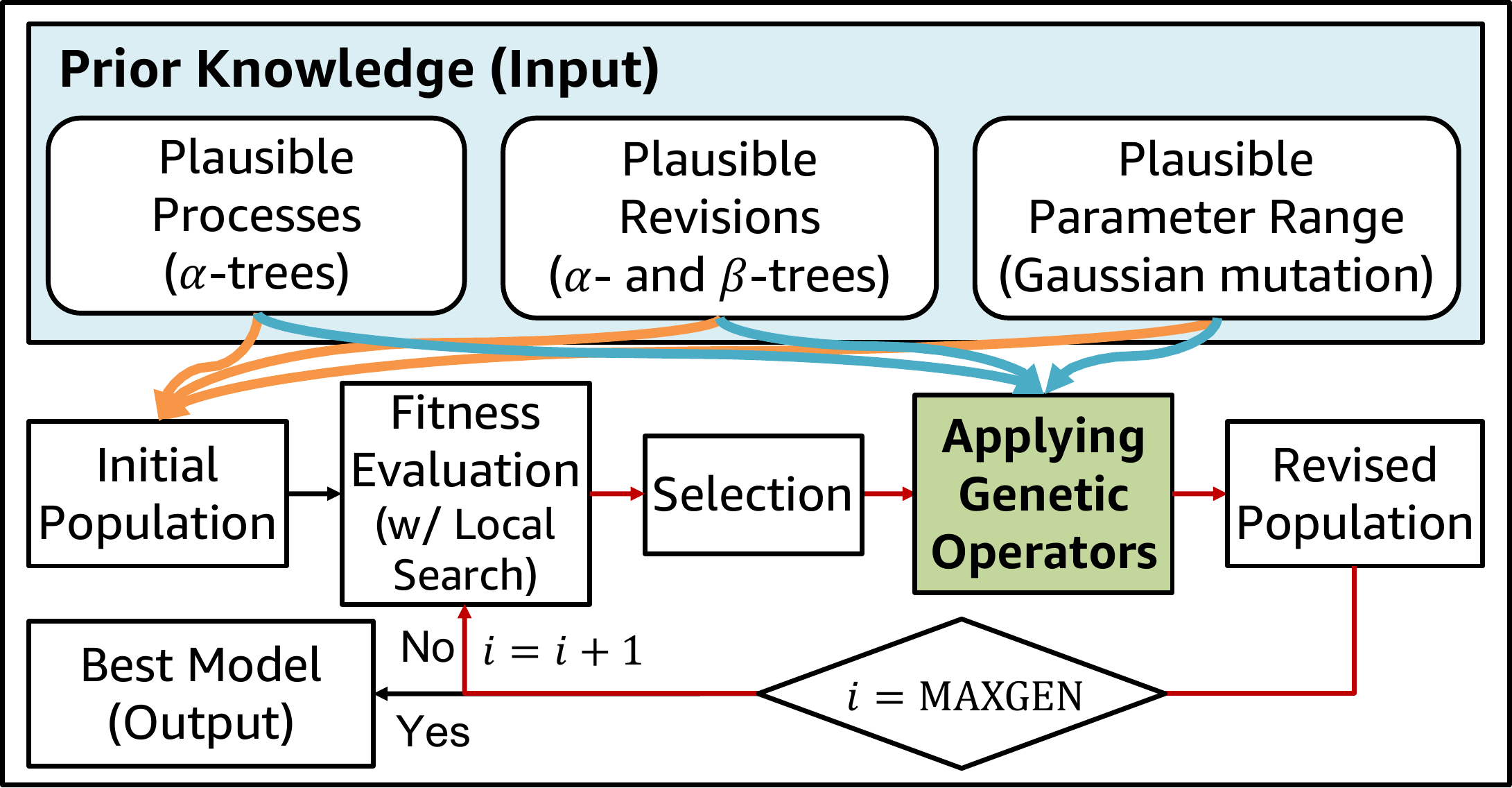}
	\caption{An overview of the model revision framework. Red loop denotes one generation.
		Prior knowledge shown in rounded boxes guides the entire model revision process.
	}
	\label{fig:framework}
	\vspace{-1em}
\end{figure}

\subsection{Knowledge-Guided Genetic Model Revision}
\label{sec:methods:framework}
In this section, we describe how genetic model revision is performed in our framework, and discuss
how we incorporate the prior domain knowledge on (1) plausible processes, (2) plausible revisions, and (3) the model parameters.

\begin{figure*}[!t]
	\centering
	\par\vspace{-1em}\par
	\captionsetup{justification=centering}
	\subcaptionbox{Before\\crossover\label{fig:operator:crossover:before}}%
	[0.13\textwidth]{\includegraphics[width=0.13\textwidth]{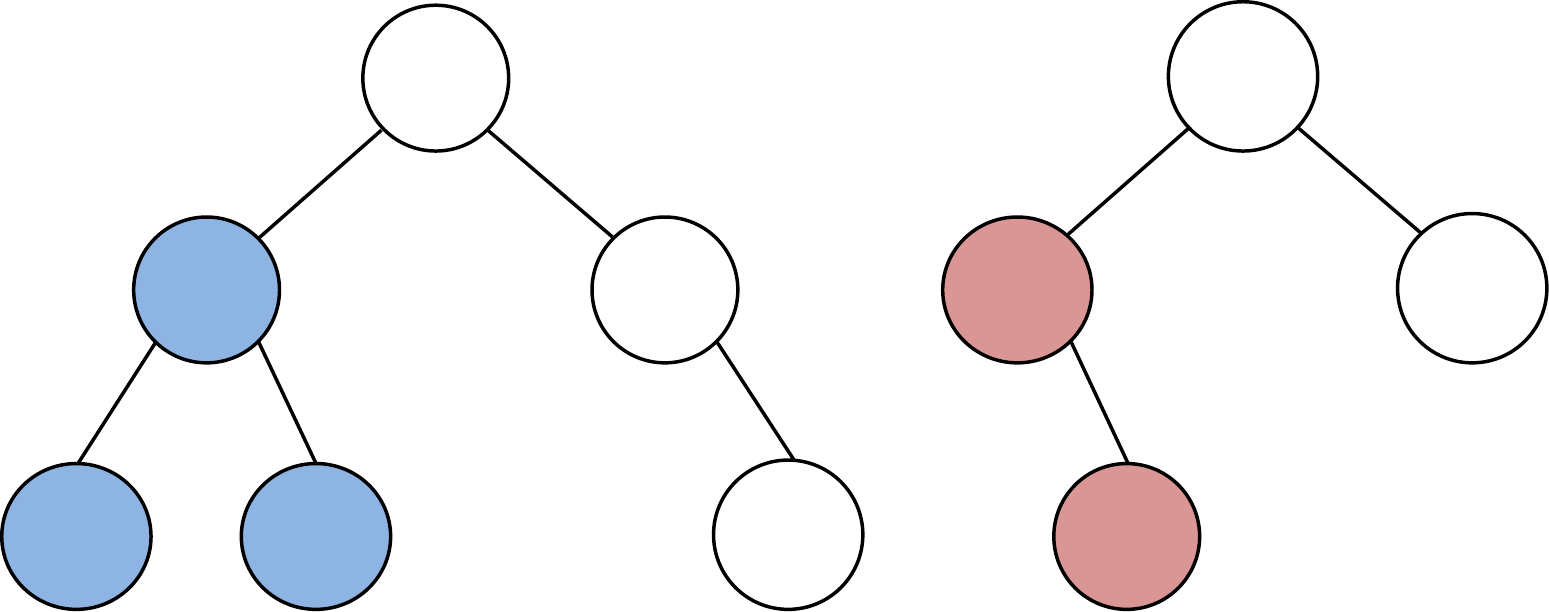}}\hfill\hfill
	\subcaptionbox{After\\crossover\label{fig:operator:crossover:after}}
	[0.13\textwidth]{\includegraphics[width=0.13\textwidth]{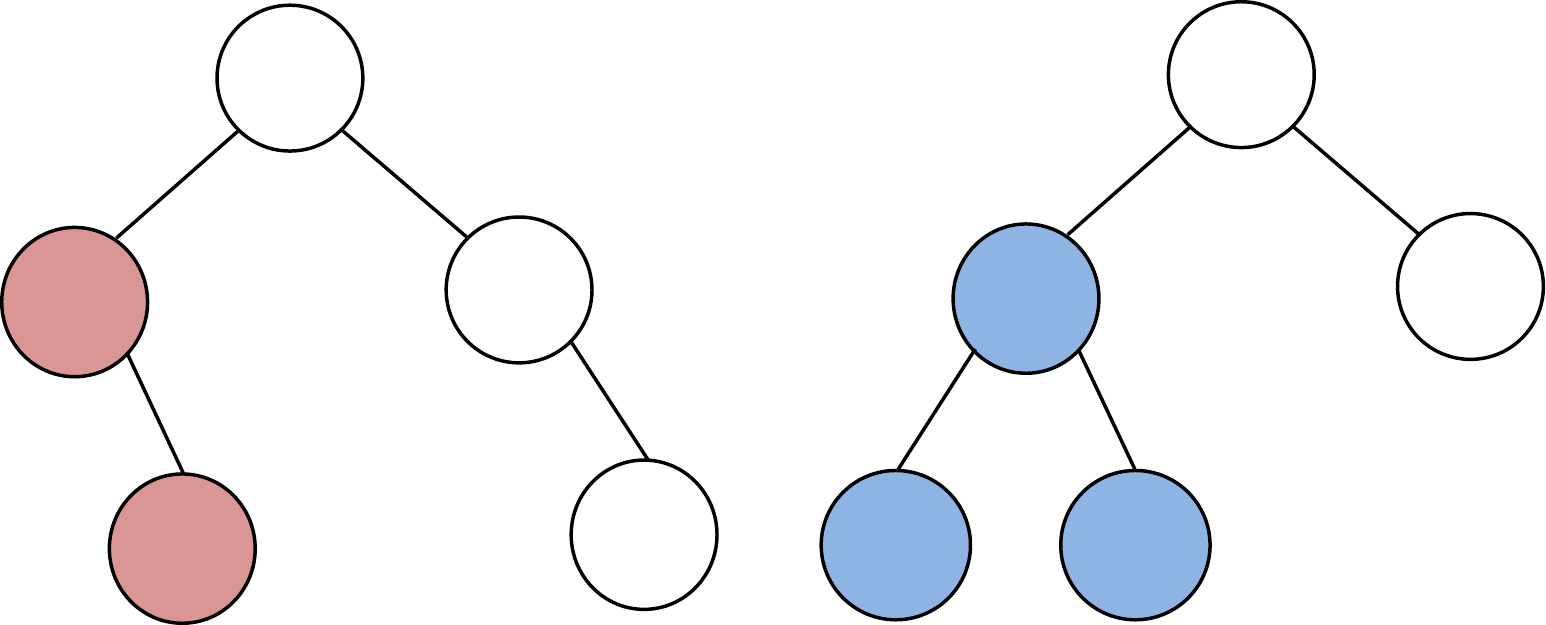}}
	\quad
	\subcaptionbox{Before subtree mutation\label{fig:operator:subtree_mutation:before}}%
	[0.12\textwidth]{\includegraphics[width=0.10\textwidth]{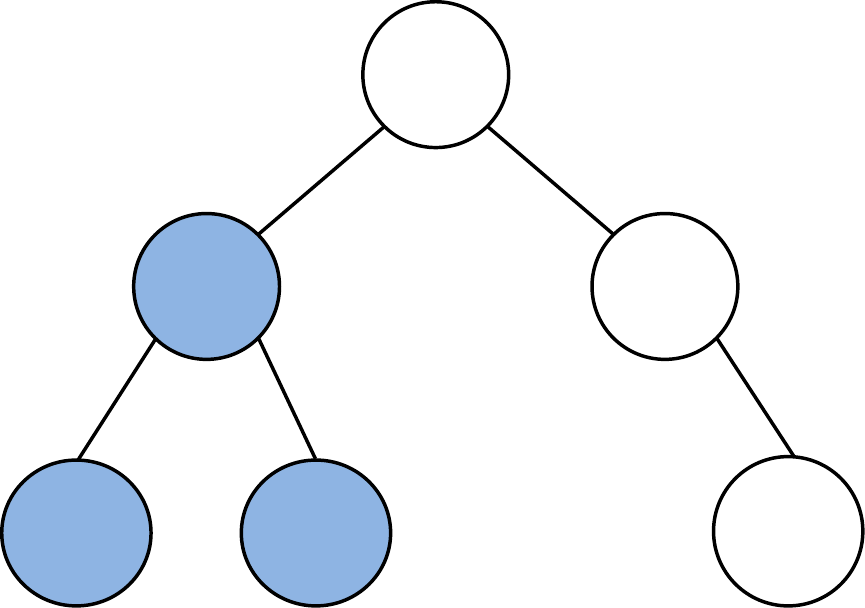}}\hfill\hfill
	\subcaptionbox{After subtree mutation\label{fig:operator:subtree_mutation:after}}%
	[0.12\textwidth]{\includegraphics[width=0.10\textwidth]{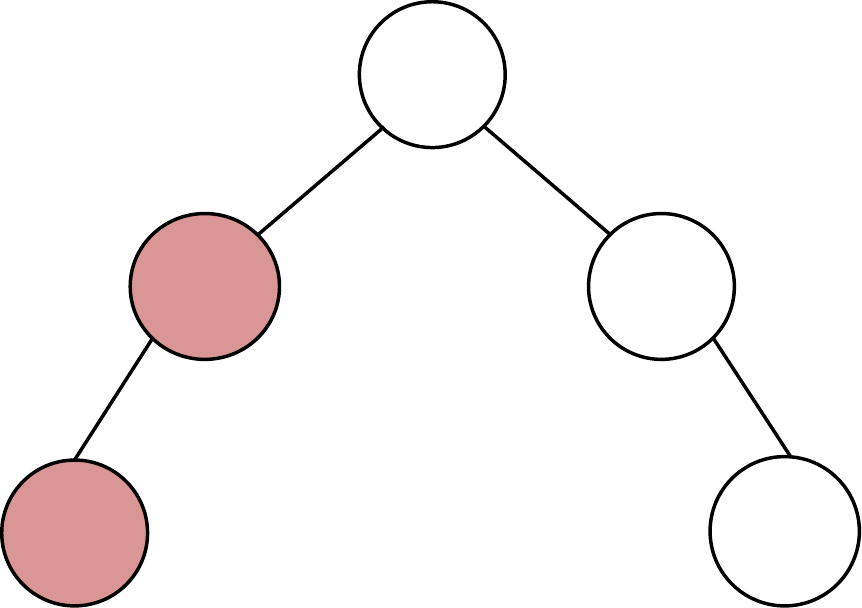}}
	\quad
	\subcaptionbox{Before point insertion\label{fig:operator:insertion:before}}%
	[0.105\textwidth]{\includegraphics[width=0.090\textwidth]{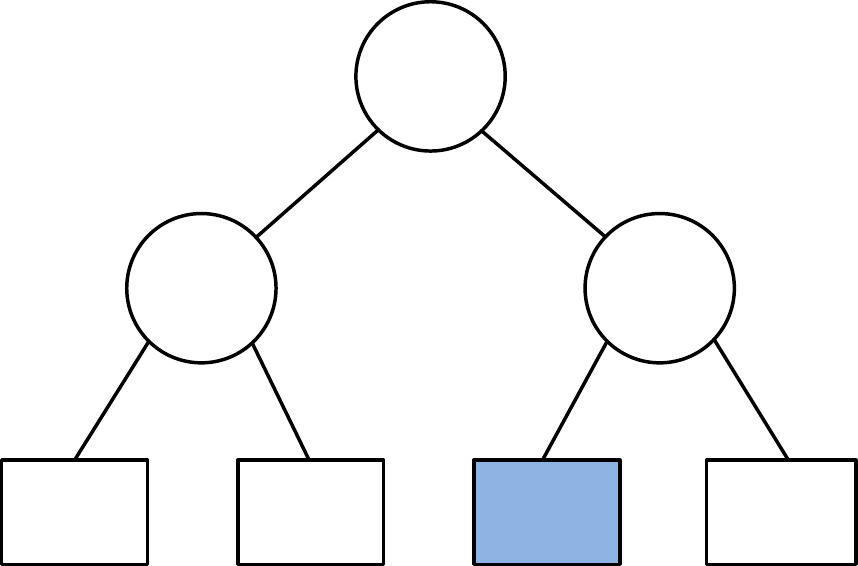}}\hfill\hfill
	\subcaptionbox{After point insertion\label{fig:operator:insertion:after}}%
	[0.105\textwidth]{\includegraphics[width=0.085\textwidth]{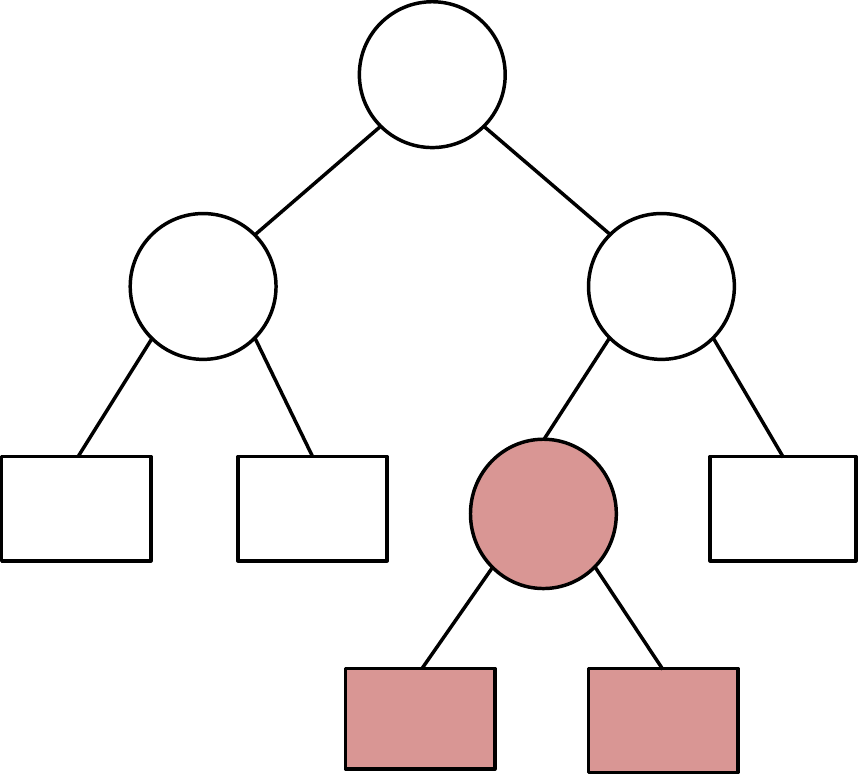}}
	\quad
	\subcaptionbox{Before point deletion\label{fig:operator:deletion:before}}%
	[0.10\textwidth]{\includegraphics[width=0.10\textwidth]{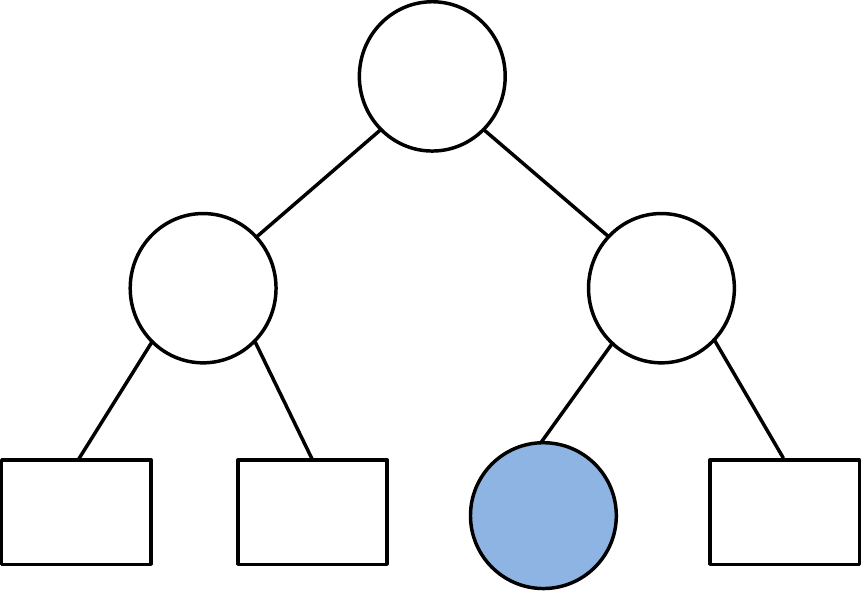}}\hfill\hfill
	\subcaptionbox{After point deletion\label{fig:operator:deletion:after}}%
	[0.10\textwidth]{\includegraphics[width=0.10\textwidth]{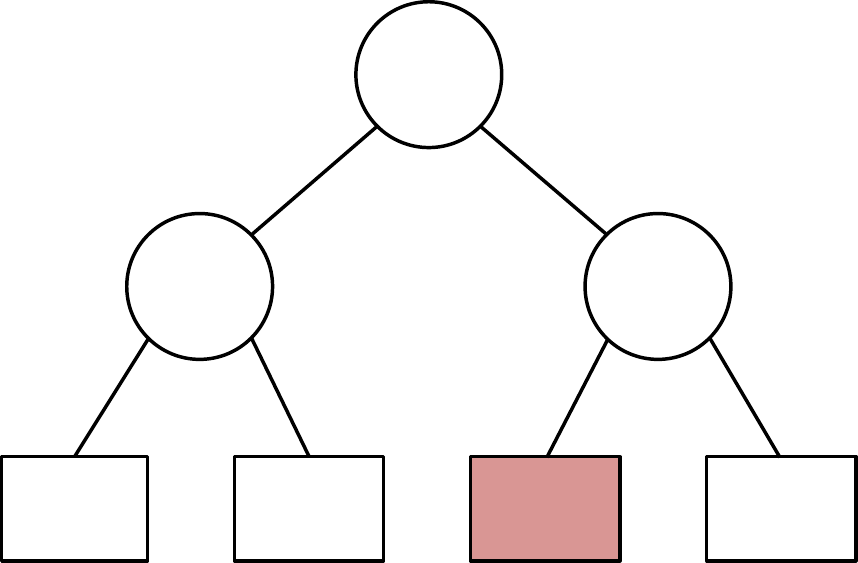}}
	\captionsetup{justification=raggedright}
	\caption{An illustration of genetic operators in TAG3P.}
	\label{fig:operator}
	\vspace{-1em}
\end{figure*}

\subsubsection{Framework Overview}
\Cref{fig:framework} shows an overview of the genetic model revision (\method) framework.
It builds upon the tree-adjoining grammar guided genetic programming (TAG3P)~\cite{DBLP:conf/eurogp/HoaiMA03}.
TAG3P is a population-based optimization algorithm 
that evolves a population of random (often unfit) initial programs (which are differential equations in our setting)
into fitter ones for a given task, over multiple generations.
TAG3P differs from standard GP in that it is a grammar-guided GP system 
where search space exploration is guided by TAG.
In~\Cref{fig:framework}, the loop marked in red corresponds to one generation.
At each generation, genetic model revision is performed on the current population
by applying genetic operators to produce a revised population of potentially fitter individuals.
Note that three types of prior knowledge (shown in rounded boxes) given to the
framework govern the entire search process, 
from population initialization to iterative model revision, until a final model is obtained.

\subsubsection{Framework Components}
Here we detail components of our TAG3P-based framework.
A detailed review of the TAG3P system can be found in~\cite{DBLP:conf/eurogp/HoaiMA03,hoaithesis2004flexible}.
\textit{Representation for a program} (called an \textit{individual}).
In our setting, a program denotes differential equations that are to be revised 
(e.g., biological process in \eqref{eq:biologicalP:raw} and \eqref{eq:biologicalZ:raw}).
In TAG3P, each program is represented as a derivation tree (\Cref{fig:dtree}).
\textit{$ \alpha $- and $ \beta $-trees.}
In \method, one $ \alpha $-tree is used to encode manually-designed minimal process as in \Cref{fig:tag_example}(a),
while other $ \alpha $-trees define model revision via substitution as in \Cref{fig:tag_example}(c).
$ \beta $-trees are defined to represent potential revisions as in \Cref{fig:tag_example}(b).
We provide more discussion on how to design these trees in~\Cref{sec:methods:priors}.

\textit{Parameters.}
Parameters include population size (\textsc{popsize}), maximum number of generations (\textsc{maxgen}), 
minimum size (\textsc{minsize}) and maximum size (\textsc{maxsize}) of individuals, 
and the probability of genetic operators.

\textit{Population Initialization.}
Individuals are created repeatedly until the population size reaches \textsc{popsize}:
TAG3P selects an individual size between \textsc{minsize} and \textsc{maxsize}, 
chooses an initial derivation tree randomly from $ \alpha $-trees, 
picks up $\beta$-trees and their adjoining addresses at random, and performs adjoining to generate an individual for the first generation.

\textit{Fitness Evaluation.}
An individual (a derivation tree) is transformed into a derived tree,
and evaluated as in standard GP.
In dynamic systems modeling, this involves evaluating revised differential equations (e.g., \Cref{fig:tag_example}(e))
for each time step, and comparing it with observed values.
More accurate individuals are given higher fitness.

\textit{Genetic Operators.}
Genetic operators make revisions to the current population to obtain others.
Among several operators in TAG3P, we introduce two representative ones (\Cref{fig:operator} illustrates these operations).

\begin{enumerate}[label=(\roman*)]
\item {Crossover}.
Two individuals are chosen by a selection mechanism, and
their subtrees are randomly selected and checked whether they are compatible.
Subtrees are compatible if each subtree can be adjoined into the node where the other subtree is attached to.
If so, the two subtrees are swapped. 
Otherwise, the previous process is retried unless the retry count has reached some predefined limit.

\item {Subtree Mutation}.
A subtree $x$ is randomly selected, and is replaced with a new subtree, 
which is of similar size to $ x $, and compatible with $ x $ (to produce a valid individual).
\end{enumerate}

\subsubsection{Incorporating Prior Knowledge}\label{sec:methods:priors}
By exploring the search space of both the structure and parameters of dynamic processes,
complex real-world systems can be modeled more accurately than can be achieved with parameter optimization alone.
However, the resulting process might be physically implausible and violate domain knowledge.
Our framework learns an accurate model that complies with prior knowledge by incorporating three types of prior knowledge.

\textbf{Prior Knowledge of Plausible Processes.}
In an effort to explain real-world phenomena, experts develop models based on domain knowledge and experience.
We harness this prior knowledge of dynamic processes, specifically 
what variables are known to be involved and how they interact with each other.
For example, the temporal dynamics of phytoplankton $ (dB_{Phy}/dt) $ in \eqref{eq:biologicalP:raw} is expressed 
as a function of zooplankton biomass $ (B_{Zoo}) $ (and other related parameters)
since zooplankton grazing pressure is known to be a major regulator of phytoplankton in a river ecosystem.
In our framework, this first type of knowledge, expressed as a differential equation, 
is encoded as an $\alpha$-tree as shown in~\Cref{fig:tag_example}(a).
Note that these input processes act as a significant knowledge transfer
at the starting point of model revision.
With classic GP systems, by contrast, we need to start from random models.

\textbf{Prior Knowledge of Plausible Revisions.}
Real-world dynamic systems are complex, often consisting of multiple intertwined processes (e.g., $ dB_{Phy}/dt $ and $ dB_{Zoo}/dt $),
each of which can be further decomposed into multiple subprocesses in a nested manner 
(e.g., $ \mu_{Phy} $ in \eqref{eq:biologicalP:raw}, which represents photosynthetic growth).
Domain experts often have an understanding of these subprocesses, such as the functionality of the subprocess, and 
plausible variables that may play a role for the subprocess.
For instance, variables that are known to affect photosynthetic growth include water alkalinity and the amount of dissolved oxygen.
Note that this subprocess corresponds to a subtree in the $\alpha$-tree representing the input process.

Our framework allows specifying variables and operations applicable for revising a specific subprocess.
This type of constraint is expressed in both $\alpha$- and $\beta$-trees.
In an $\alpha$-tree, extensible subtrees are placed under a special node, 
whose name starts with ``Ext'', denoting a revision that can be made to the corresponding subtree via tree-adjoining.
In an equation form, we denote a subprocess $ f(\cdot) $ extensible via ``Ext'' by $ \{f(\cdot)\}\!\odot\!\text{Ext}$.
We then generate a list of $ \beta $-trees for each combination of variables and operators,
which have the corresponding ``Ext'' as the root node, and a foot node of the same type
to define allowable operations for the given subtree.

Importantly, we distinguish between the operators that are applied directly to the initial process (called \textit{connectors}), and 
those that are applied to the subprocesses, which extend the initial process, but do not belong to the initial process (called \textit{extenders}).
This is to preserve the initial process by applying a limited set of operations to it,
while giving a greater freedom for extenders to make improvements to the initial process.

\begin{figure}[t!]
	\centering
	\makebox[0.5\linewidth][c]{\includegraphics[trim={0 0 0 0},clip,width=1.075\linewidth]{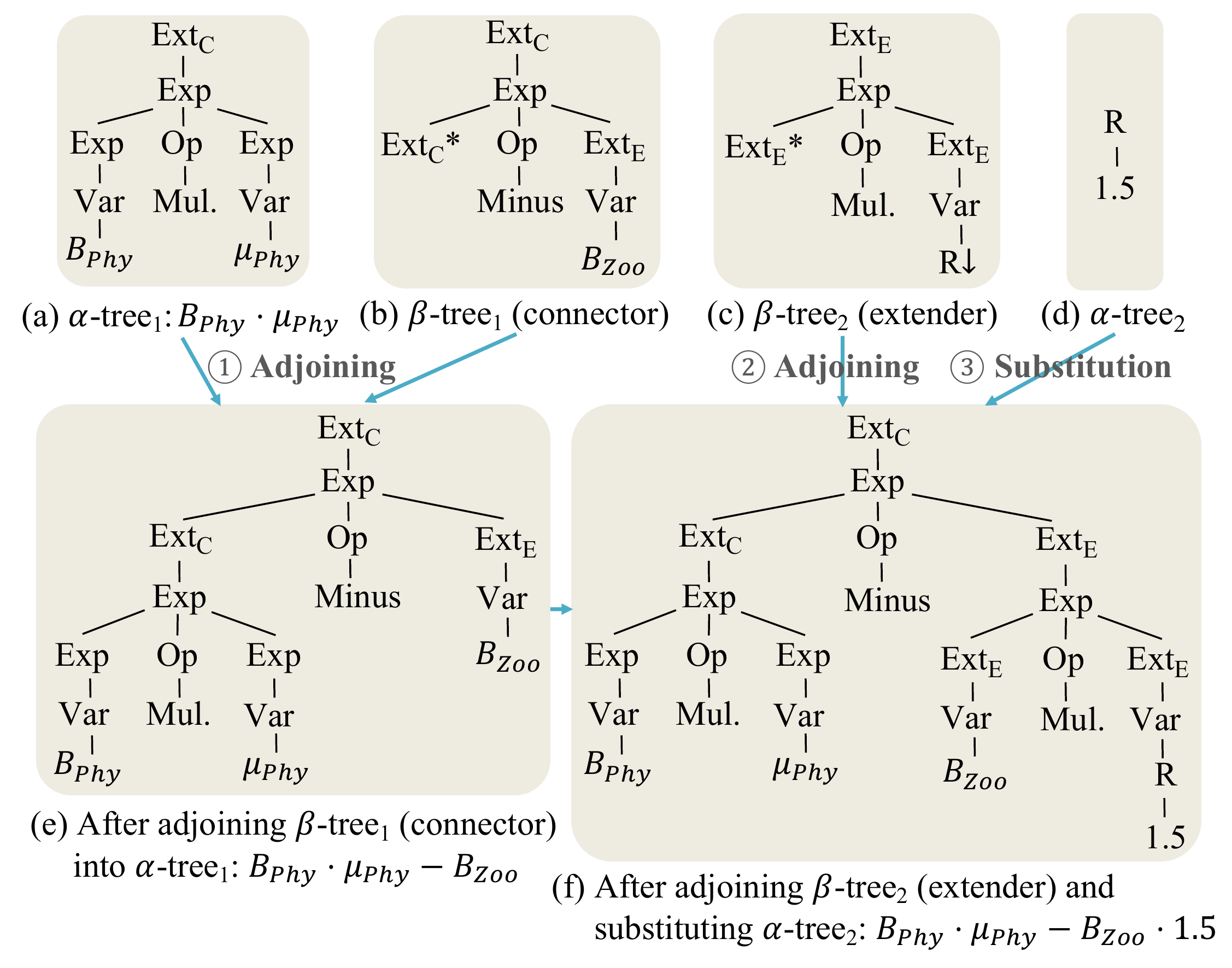}}
	\setlength{\abovecaptionskip}{0pt}
	\caption{(a)--(d): $ \alpha $- and $ \beta $-trees for (a) the initial process, 
		revision via adjoining with (b) a connector and (c) an extender, and (d) substitution. 
		(e), (f): Resulting trees after revision via a connector and an extender (see text for details).}
	\label{fig:model_revision_example}
\end{figure}

As an example, consider again this simplified process %
$ d {B_{\mathit{Phy}}} / dt = \{ B_\mathit{Phy} \cdot {\mu}_\mathit{Phy} \} \odot \text{Ext} $, extensible via Ext.
This can be encoded as an $ \alpha $-tree in \Cref{fig:model_revision_example}(a).
Note that Ext\textsubscript{c} (denoting connectors) is used for the root node.
We assume that we have two variables $ B_{Zoo} $ and $ R $,
deduction (Minus) as a connector and multiplication (Mul.) as an extender.
\Cref{fig:model_revision_example}(b) and (c) show two $ \beta $-trees that can be generated 
(Ext\textsubscript{c} and Ext\textsubscript{E} denoting a connector and an extender).
Adjoining \Cref{fig:model_revision_example}(a) and (b) produces $ B_\mathit{Phy} \cdot {\mu}_\mathit{Phy} - B_\mathit{Zoo} $ in \Cref{fig:model_revision_example}(e);
subsequently adjoining \Cref{fig:model_revision_example}(c) and substituting \Cref{fig:model_revision_example}(d) into \Cref{fig:model_revision_example}(e) 
yields a revised process $ B_\mathit{Phy} \cdot {\mu}_\mathit{Phy} - B_\mathit{Zoo} \cdot 1.5 $ in~\Cref{fig:model_revision_example}(f)
(extensions removed for brevity).
Note that since connectors and extenders use different symbols (Ext\textsubscript{c} and Ext\textsubscript{e}),
connector $ \beta $-trees cannot adjoin into $ \alpha $-tree nodes with extender symbols, and vice versa.

\textbf{Prior Knowledge about Model Parameters.}
From previous research and experience on dynamical systems,
domain experts often have the information on the plausible distributions of the model parameters.
Even if we obtain a highly accurate model, if its parameters are not within a realistic range,
that model is not considered a good representation of underlying processes.
In \method, the domain knowledge of model parameters is summarized as 
the expected value and allowed range of parameter values.
For effective search, ranges need to be chosen to cover most practically feasible values.
We assume that naturally occurring values follow a truncated Gaussian distribution centered around the expected value.

To optimize model parameters, we apply a genetic operator, \textit{Gaussian mutation},
which locates all constant parameters in an individual,
and updates them to new values sampled from their associated Gaussian distribution.
In the beginning, parameters are set to the expected value.
When Gaussian mutation is applied to a parameter, a new value is generated, and
it becomes the new mean of the Gaussian distribution for that parameter.
If the sampled value lies outside of the given range, the boundary value is used instead.

For river water quality modeling, we initially set the standard deviation to $1/4$ of the parameter mean,
as that covers the range of most observable parameter values. 
We then ramp down the standard deviation linearly in the final $ k $ generations so that it becomes smaller in later generations.

\subsection{Applying \method to Real-World Problems}
\label{sec:methods:rivermodeling}

\begin{table}[!t]
	\centering
	\setlength{\tabcolsep}{1.0pt}
	\setlength{\abovecaptionskip}{0pt}
	\caption{Variables, connectors, and extenders used by extensions. 
		$ R $ denotes a random variable between 0 and 1.}
	\begin{tabular}{c|r|c|r|c|r} 
		\toprule
		Extension & \makecell{Variables} & Extension & \makecell{Variables} & Extension & \makecell{Variables} \\ \midrule
		Ext1 & $ ~V_{cd}, V_{ph}, V_{alk}, R~$ & Ext5 & $ ~V_{tmp}, R~ $ & Ext8 & $~V_{tmp}, R$ \\ 
		Ext2 & $ ~V_{sd}, R~$ & Ext6 & $ ~V_{tmp}, R~ $ & Ext9 & $ ~V_{tmp}, R $ \\ 
		Ext3 & $ ~V_{do}, V_{ph}, V_{alk}, R~$ & Ext7 & $ ~V_{tmp}, R~$ & & \\ \bottomrule \toprule
		Connectors & \multicolumn{5}{r}{$ + $ for extensions 1--3, $ \times $ for extensions 5--9} \\
		Extenders & \multicolumn{5}{r}{$ +, -, \times, \div, \log, \exp $ for all extensions} \\
		\bottomrule
	\end{tabular}
	\vspace{-1em}
	\label{tab:extension}
\end{table}

\textbf{River Water Quality Modeling.}
We show how the proposed framework can be applied to the modeling of river water quality introduced in \Cref{sec:prelim},
in which the goal is to make an accurate temporal %
prediction of the phytoplankton biomass $(B_{Phy})$.
We capture the expert knowledge on the dynamics of phytoplankton ($ dB_{Phy} / dt $) and zooplankton ($ dB_{Zoo} / dt $), and 
plausible revisions as follows. %

{%
\setlength{\abovedisplayskip}{3pt}
\setlength{\belowdisplayskip}{\abovedisplayskip}
\setlength{\abovedisplayshortskip}{0pt}
\setlength{\belowdisplayshortskip}{3pt}
\footnotesize
\begin{align}
\label{eq:biologicalP}
\frac{d {B_{\mathit{Phy}}}}{d t}	& = \left\{B_\mathit{Phy} \cdot ({\mu}_\mathit{Phy} - {\gamma}_\mathit{Phy}) - B_\mathit{Zoo} \cdot \varphi \right\}\!\odot \!\text{Ext1}\\
\mu_{\mathit{Phy}}	& = \left\{C_\mathit{UA} \cdot f(V_\mathit{lgt}) \cdot g(V_{n},V_{p},V_{si}) \cdot h(V_\mathit{tmp})\right\}\!\odot\!\text{Ext3} \nonumber\\
\gamma_{\mathit{Phy}}	& = \left\{C_{\mathit{BRA}} \right\}\!\odot\!\text{Ext5} \nonumber\\
\varphi		& = \left\{C_\mathit{MFR} \cdot \mathit{\lambda_{Phy}} \right\}\!\odot\!\text{Ext6} \nonumber\\ 
\mathit{\lambda_{Phy}}	& = ({B_\mathit{Phy}}-C_\mathit{Fmin}) / (C_\mathit{FS}+B_\mathit{Phy} - C_\mathit{Fmin}) \nonumber\\
f(V_{\mathit{lgt}})	& = (V_{\mathit{lgt}} / C_{\mathit{BL}}) \cdot e^{1-(V_{\mathit{lgt}} / C_{\mathit{BL}})} \nonumber\\
g(V_{n},V_{p},V_{\mathit{si}})	& =\min\left( V_{n} / (C_{N}+V_{n}), V_{p} / (C_{P}+V_{p}), V_{si} / (C_{SI}+V_{si}) \right) \nonumber\\
h(\mathit{V_{tmp}})	& = \max( e^{-C_{\mathit{PT}} (V_{\mathit{tmp}} - C_{\mathit{BTP1}})^2}, e^{-C_{\mathit{PT}} (V_{\mathit{tmp}} - C_{\mathit{BTP2}})^2} ) \nonumber\\[0.5em]
\label{eq:biologicalZ}
\frac{d {B_{\mathit{Zoo}}}}{d t}	& = \left\{{B_{\mathit{Zoo}}} \cdot ({\mu}_{\mathit{Zoo}} - {\gamma}_\mathit{Zoo} - {\delta}_\mathit{Zoo} )\right\}\!\odot\!\text{Ext2} \\ 
{\mu}_\mathit{Zoo}	& = \left\{ C_\mathit{UZ} \cdot \mathit{\lambda_{Phy}} \right\}\!\odot\!\text{Ext7} \nonumber\\
{\gamma}_\mathit{Zoo}	& = \left\{ C_\mathit{BRZ} \right\}\!\odot\!\text{Ext8} + C_\mathit{BMT} \cdot \varphi \nonumber\\
{\delta}_\mathit{Zoo}	& = \left\{ C_\mathit{DZ} \right\}\!\odot\!\text{Ext9} \nonumber
\end{align}}%
\Cref{tab:extension} presents the list of variables, connectors, and extenders applicable to each extension.
$ R $ denotes a variable that is randomly initialized.
Constant parameters (those starting with $ C $) are initialized and updated via Gaussian mutation,
based on their mean and exploration range given in~\Cref{tab:process_constants}.

These extensions are defined based on an extensive river modeling experience of a freshwater ecologist,
denoting different types of extensions plausible for specific subprocesses.
For example, electric conductivity ($ V_{cd} $) applies to the dynamics of phytoplankton via Ext1, but not to that of zooplankton.

\textbf{Revising Multiple Processes.} While input processes are to be represented using one $ \alpha $-tree, 
here we have two differential equations \eqref{eq:biologicalP} and \eqref{eq:biologicalZ}.
Multiple equations can be encoded as a single $ \alpha $-tree
by first representing each equation in separate trees, and then combining them into one $ \alpha $-tree under a new, common root node.
Then this combined $ \alpha $-tree can be evolved in the same manner as in simpler cases,
and decomposed into multiple equations when performing fitness evaluation.

\textbf{Application to Other Problems.}
\method provides general mechanisms to represent and revise process equations guided by prior knowledge,
so is readily applicable to diverse problem settings.
The only problem-dependent component %
is representing domain-specific knowledge as discussed in \Cref{sec:methods:priors};
this itself is a general technique applicable to various problems.

\begin{table}[t!]
\setlength{\tabcolsep}{1.0pt}
\setlength{\aboverulesep}{0pt}
\setlength{\belowrulesep}{0pt}
\centering
\caption{Constant parameters that are updated via Gaussian mutation. 
	Prior knowledge is captured as the mean and exploration bounds (minimum and maximum values).}
\makebox[0.4\textwidth][c]{
	\begin{tabular}{l|l|r|r|r|r} \toprule
		& \makecell[c]{Description} & \makecell[c]{Mean} & \makecell[c]{Min} & \makecell[c]{Max} & \makecell[c]{Unit}\\ \midrule
		$\mathit{C_{UA}}$ & Max growth rate of phytoplankton & 1.89 & 0.1 & 4.0 & day$^{-1}$ \\
		$\mathit{C_{UZ}}$ & Max growth rate of zooplankton & 0.15 & 0.0 & 0.3 & day$^{-1}$ \\
		$\mathit{C_{BRA}}$ & Breath rate of phytoplankton & 0.021 & 0.0 & 0.17 & day$^{-1}$ \\
		$\mathit{C_{BRZ}}$ & Breath rate of zooplankton & 0.05 & 0.0 & 0.2 & day$^{-1}$ \\
		$\mathit{C_{MFR}}$ & Maximum feeding rate & 0.19 & 0.01 & 0.8 & day$^{-1}$ \\
		$\mathit{C_{DZ}}$ & Death rate of zooplankton & 0.04 & 0.01 & 0.1 & day$^{-1}$ \\
		$\mathit{C_{FS}}$ & Half-saturation constant of food & 5.0 & 4.0 & 6.0 & $\upmu$g L$^{-1}$ \\
		$\mathit{C_{BTP1}}$ & Blue-green optimal temperature & 27.0 & 20.0 & 34.0 & $^\circ$C \\
		$\mathit{C_{BTP2}}$ & Diatom optimal temperature & 5.0 & 1.0 & 20 & $^\circ$C \\
		$\mathit{C_{Fmin}}$ & Minimum food concentration & 1.0 & 0.1 & 1.9 & $\upmu$g L$^{-1}$ \\
		$\mathit{C_{BL}}$ & Best light for phytoplankton & 26.78 & 24.0 & 30.0 & \makecell{MJ\\ m$^{-2}$ d$^{-1}$} \\
		$\mathit{C_{N}}$ & Half-saturation constant of nitrogen & 0.0351 & 0.02 & 0.05 & mg L$^{-1}$ \\
		$\mathit{C_{P}}$ & Half-saturation constant of phosphorus & 0.00167 & 0.001 & 0.02 & mg L$^{-1}$ \\
		$\mathit{C_{SI}}$ & Half-saturation constant of silica & 0.00467 & 0.001 & 0.2 & mg L$^{-1}$ \\
		$\mathit{C_{BMT}}$ & Breath multiplier on grazing & 0.04 & 0.01 & 0.07 & N/A \\
		$\mathit{C_{PT}}$ & \makecell[l]{Temperature coefficient for\\ phytoplankton growth} & 0.005 & 0.003 & 0.2 & $^\circ$C$^{-2}$ \\
		$\varphi$ & Grazing rate of zooplankton & N/A & N/A & N/A & d$^{-1}$ \\ \bottomrule
	\end{tabular}
}
\label{tab:process_constants}
\end{table}

\begin{table}[t!]
\centering
\setlength{\tabcolsep}{4pt}
\setlength{\aboverulesep}{0pt}
\setlength{\belowrulesep}{0pt}
\caption{Temporal variable parameters in the river process.}
\makebox[0.4\textwidth][c]{
	\begin{tabular}{l|l|l|l} \toprule
		Parameter & Description & Parameter & Description\\ \midrule
		$V_{lgt}$ & Irradiance (light intensity) & $V_{do}$ & Dissolved oxygen \\
		$V_{n}$ & Nitrogen concentration & $V_{cd}$ & Electric conductivity \\
		$V_{p}$ & Phosphorus concentration & $V_{ph}$ & pH \\
		$V_{si}$ & Silica concentration & $V_{alk}$ & Alkalinity \\
		$V_{tmp}$ & Water temperature & $V_{sd}$ & Water transparency \\ \bottomrule
	\end{tabular}
}
\label{tab:process_variables}
\vspace{-1em}
\end{table}

\subsection{Improving the Efficiency and Effectiveness}
\label{sec:methods:improving}

For efficient and effective optimization, 
we apply three orthogonal speedup techniques, together with local search.

\textbf{Evaluation Short-Circuiting.}
Modeling temporal processes involves incremental fitness evaluation over a period of time,
in which case intermediate fitness may provide a reasonable estimate of the final fitness.
Also, GP is known to be robust to noisy evaluation.
Based on these observations, fitness evaluation can be short-circuited, using the estimate as a surrogate of the final fitness.
Early termination is such an approach where fitness evaluation is stopped 
when the intermediate fitness gets worse than the previous best fitness obtained from full evaluations.
In \method, we design a generalized evaluation short-circuiting technique (\Cref{alg:evalshortcircuiting}),
which allows controlling the eagerness of early termination (via the \textit{threshold} parameter) and use of different extrapolation methods.

\textbf{Runtime Compilation.}
A tree representing temporal processes needs to be evaluated multiple times over some time period, and
each such evaluation can be done by recursively evaluating subtrees, providing the model parameter values appropriately at each step.
Instead, we use runtime compilation, which enables more efficient evaluation than repeated tree parsing:
a program encoded in the tree is converted into the corresponding source code, 
compiled at runtime, and dynamically loaded to be used for fitness evaluation.

\textbf{Tree Caching.}
We cache the results of tree evaluation, and reuse them when we need to reevaluate the same trees.
By using additional memory to store evaluation results, we avoid redundant computations.
Note that the effectiveness of caching depends on the hit rate.
\method improves the hit rate by 
algebraically simplifying the trees before they are evaluated.
We show the benefits of speedup techniques in \Cref{sec:exp:speedup}.

\textbf{Local Search.} Local search aims to improve the search effectiveness by 
making incremental, local revisions to an individual.
We apply two local search operators, \textit{insertion} and \textit{deletion}.
Insertion randomly chooses an open adjoining address of a TAG derivation tree, 
and adds a randomly selected compatible auxiliary tree to the chosen location.
Deletion removes a random node from the derivation tree.
\Cref{fig:operator} illustrates these operations.
Specifically, we perform stochastic hill-climbing local search, where a tree resulting from crossover and mutation goes through 
a series of local search, applying insertion and deletion with equal probability, and adopting the change if it improves the fitness.

\begin{algorithm}[!t]
\small
\DontPrintSemicolon
\SetKwComment{Comment}{$\triangleright$\ }{}
\KwIn{\textit{ind} (an individual to be evaluated), \textit{threshold} (a non-negative value that determines when to check for evaluation short-circuiting), \textit{numFitcases} (number of fitness cases),
\textsc{Extrapolate} (a function to extrapolate intermediate fitness).
}
\KwOut{\textit{fitness} (evaluated fitness of the given individual \textit{ind}).}

\SetKwFunction{FMain}{{\normalfont \textsc{FitnessEvaluation}}}
\SetKwProg{Fn}{Function}{\normalfont :}{}
\Fn{\FMain{\textit{ind}, \textit{threshold}, \textit{numFitcases}}}{
	$\textit{bestPrevFull} \gets \infty$\\
	$\textit{fitness} \gets 0$\\
	$ \textit{i} \gets 0 $\\
	\While{$ \textit{i} < \textit{numFitcases} $}{
		Update $\textit{fitness}$ of $\textit{ind}$ using fitness case $ i $\\
		\If{$ \textit{fitness} > \textit{bestPrevFull} \times \textit{threshold} $}{
			\textit{estFitness} $ \gets $ \textsc{Extrapolate}(\textit{fitness}, $ \textit{i} $, \textit{numFitcases})
			
			\If{$\textit{estFitness} > \textit{bestPrevFull}$}{
				\Return{\textit{estFitness}}\Comment*[r]{\normalfont Short Circuiting}
			}
		}
		$ \textit{i} \gets \textit{i} + 1 $
	}
	\If{$ \textit{fitness} < \textit{bestPrevFull} $}{
		$ \textit{bestPrevFull} \gets \textit{fitness} $
	}
	\Return{\textit{fitness}}\Comment*[r]{\normalfont Full Evaluation}
}
\caption{{Evaluation Short-Circuiting}}
\label{alg:evalshortcircuiting}
\end{algorithm}

%% file: 040experiments.tex
In this section, we evaluate our \method framework 
via a case study on river modeling.
We address the following questions.

\begin{enumerate}[label=\textbf{Q{{\arabic*}}.},ref=Q\arabic*]
	\item \label{sec:exp:q1} \textbf{Prediction Accuracy:} 
	How accurately does the \method framework forecast river water quality?

	\item \label{sec:exp:q2} \textbf{Ecological Analysis:}
	How does \method revise the input process, and does the revision make sense from an ecological standpoint? 
	Which variables are important in the revision?

	\item \label{sec:exp:q3} \textbf{Speedup Techniques:}
	How much do the speedup techniques improve efficiency?

\end{enumerate}

\subsection{Dataset and Modeling Task Description}
\label{sec:exp:dataset}

\begin{figure}[t!]
	\par\vspace{-0.5em}\par
	\centering
	\includegraphics[width=0.80\linewidth]{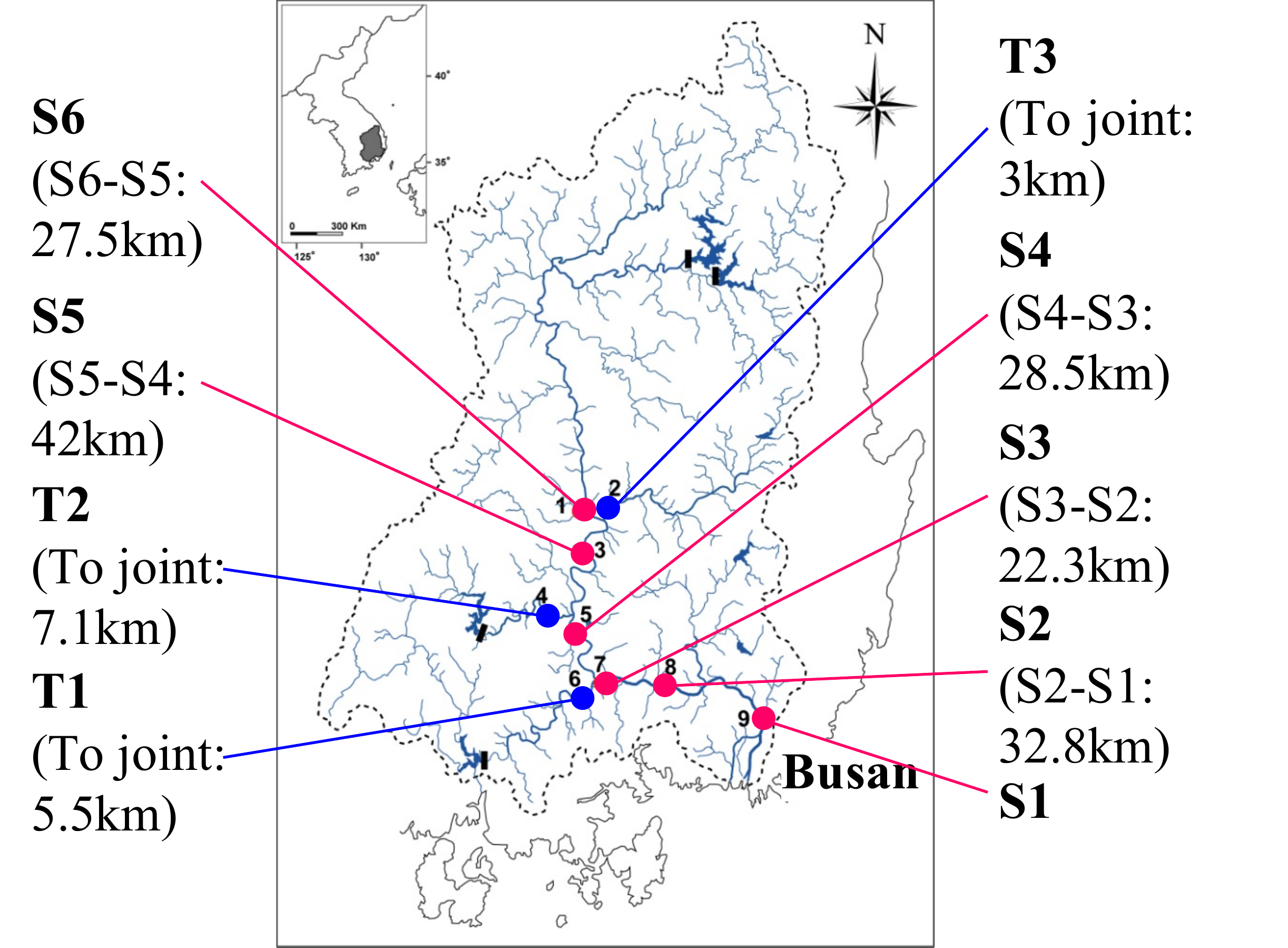}
	\caption{The Nakdong River basin in South Korea. Circles denote measuring stations. 
		Red circles (S1--S6) are on the main channel, while blue ones (T1--T3) are on the major tributaries.
	}
	\label{fig:studysite}
	\vspace{-1em}
\end{figure}

The Nakdong River catchment in South Korea is one of the largest water-quality monitoring networks supporting long-term ecological research.
Our dataset is a collection of measurements for 13 years (1996--2008) at nine stations located in the catchment; (\Cref{fig:studysite}):
six (S1--S6) are sited on the main channel, while three (T1--T3) are on major tributaries.
The dataset contains five types of variables: 
geographical (e.g., catchment area), hydrological (e.g., flow rate), meteorological (e.g., irradiance), 
physicochemical (e.g., water temperature), and biological (e.g., chlorophyll a).
Most were measured daily, except for
nutrient concentrations and chlorophyll a, which were
measured weekly (at S1) or bi-weekly (at others).
For those variables measured with a longer interval,
we performed linear interpolation to obtain values between measurements.
Note that these measurements provide values 
for the temporal variables in the river process (i.e., those starting with $ V $ in \eqref{eq:biologicalP:raw}, \eqref{eq:biologicalZ:raw} and \Cref{tab:extension}).
\Cref{tab:process_variables} gives a description of the temporal variables.
Given these measurements, our goal is to forecast the algal biomass at the lowest station (S1)
due to its geographical importance (around ten million people live near S1).
Specifically, we make knowledge-guided revisions to 
the \textit{biological} process \eqref{eq:biologicalP:raw} and \eqref{eq:biologicalZ:raw} 
to closely estimate observed algal biomass at S1.

In addition to the biological process, 
river modeling involves modeling the flow of water bodies in the river (called the \textit{hydrological} process).
What this hydrological process models includes 
how water bodies are discharged from stations, how the rainfall is absorbed into the river, etc.
As our focus is on improving the biological process, we use a known hydrological process
\Cref{sec:rivermodeling} gives details of the hydrological process, and 
how a river system with multiple stations is implemented.

\subsection{Comparators}
\label{sec:exp:baseline}

For evaluation, we use representative comparators in the four classes of methods for modeling dynamic systems.

\subsubsection{Knowledge-Driven Modeling}
(a) \manual. This is the biological process 
in~\eqref{eq:biologicalP:raw} and \eqref{eq:biologicalZ:raw}, designed by domain experts.

\subsubsection{Data-Driven Modeling}
(a) \rnn (Recurrent Neural Network).
We use long short-term memory (LSTM),
predicting the phytoplankton biomass at S1 at the next time step from observed variables at the current time.
We experiment with two variants:
{\rnnbusan} uses variables observed at station S1 alone;
{\rnnall} uses variables observed at all nine stations.
(b) {\arimax} is widely used for time series forecasting.
As with \rnn, we consider two variants differing in variables used
(denoted as {\arimaxBusan} and {\arimaxAll}).

\subsubsection{Model Calibration}
Given the biological process in~\eqref{eq:biologicalP:raw} and \eqref{eq:biologicalZ:raw},
model calibration methods optimize the values of process parameters without revising the form of equations.
We use the following widely-used approaches:
(a) \ga (genetic algorithm)
(b) \mc (Monte Carlo)
(c) \lhs (Latin hypercube sampling)
(d) \mle (maximum likelihood estimation)
(e) \mcmc (Markov chain Monte Carlo)
(f) \sa (simulated annealing)
(g) \dream (differential evolution adaptive metropolis~\cite{vrugt2016markov})
(h) \sceua (shuffled complex evolution~\cite{duan1994optimal})
(i) \demcz (differential evolution Markov chain~\cite{vrugt2008treatment}).

\subsubsection{Model Revision}
(a) \gggp (grammar guided GP). We perform model revision using \gggp; that is, 
\gggp receives the biological process in~\eqref{eq:biologicalP:raw} and \eqref{eq:biologicalZ:raw} as input, 
and updates both the model structure and parameter values.

\Cref{sec:exp:settings} provides experimental settings of all methods.

\subsection{Performance Evaluation}
\label{sec:exp:evaluation}

We use RMSE (root mean square error) and MAE (mean absolute error).
Let $ y_t $ and $ \hat{y}_t $ denote the observed and predicted values at time $ t $, respectively.
Given observations for $ T $ time steps,
we have $ \mathbf{y} = (y_1, \ldots, y_T) $ and $ \mathbf{\hat{y}} = (\hat{y}_1, \ldots, \hat{y}_T) $.
RMSE is a quadratic score which measures the average magnitude of prediction errors, giving a relatively large weight to large errors. 
RMSE is defined by
$\sqrt{\frac{1}{T}\!\sum_{t=1}^T (\hat{y}_t\!-\!y_t)^2}$.
MAE is a linear scoring rule for measuring the average magnitude of prediction errors, giving equal weights to individual differences.
MAE is defined by $ \frac{1}{T}\!\sum_{t=1}^T | \hat{y}_t\!-\!y_t | $.
Both RMSE and MAE range from zero to $ \infty $, and lower values indicate a better prediction.

\textbf{Fitness Function.} RMSE is used as a fitness function.

\begin{table}[!t]
	\setlength{\tabcolsep}{0.6em}
	\centering
	\setlength{\aboverulesep}{0pt}
	\setlength{\belowrulesep}{0pt}
	\setlength{\extrarowheight}{2pt}
	\setlength{\abovecaptionskip}{5pt}
	\caption{\method achieves the best forecasting accuracy (7\% and 13\% more accurate than the second best method in terms of RMSE and MAE, respectively), among a variety of methods. Best results are underlined.}
	\makebox[0.4\textwidth][c]{
		\begin{tabular}{c|c|r|r|r|r} 
			\toprule
			\multirow{2}{*}{\makecell[c]{Method\\Class}} & \multirow{2}{*}{Method} & \multicolumn{2}{r|}{\makecell[c]{Training (96--05)}} & \multicolumn{2}{r}{\makecell[c]{Test (06--08)}~~} \\ 
			& & \makecell[c]{RMSE} & \makecell[c]{MAE} & \makecell[c]{RMSE} & \makecell[c]{MAE} \\ \midrule
			\makecell[c]{Knowledge-\\driven} & \manual & 2.79e+9 & 2.15e+8 & 2.23e+6 & 7.93e+5 \\ \midrule
			\multirow{4}{*}{\makecell[c]{Data-\\driven}} & \rnnbusan & {19.605} & 11.533 & {23.057} & 16.833 \\
			& \rnnall & 21.326 & 13.166 & 23.009 & 16.276 \\
			& \arimaxBusan & 12.710 & \dunderline{0.8pt}{5.012} & 37.770 & 25.504 \\ 
			& \arimaxAll & \dunderline{0.8pt}{12.365} & 5.775 & 260.468 & 71.471 \\ \midrule
			\multirow{8}{*}{\makecell[c]{Model\\calibration}} & \ga & 26.329 & 14.693 & 20.308 & 13.291 \\ 
			& \mc & 26.581 & 14.426 & 19.259 & 12.675 \\ 
			& \lhs & 26.812 & 14.536 & 18.287 & 12.064 \\ 
			& \mle & 26.033 & 14.408 & 19.513 & 13.242 \\ 
			& \mcmc & 26.514 & 14.554 & 18.661 & 12.480 \\ 
			& \sa & 26.463 & 14.585 & 18.740 & 12.532 \\ 
			& \dream & 26.825 & 14.853 & 19.281 & 12.581 \\ 
			& \sceua & 25.995 & 14.353 & 19.876 & 13.275 \\ 
			& \demcz & 26.227 & 14.432 & 18.904 & 12.869 \\ \midrule
			\multirow{2}{*}{\makecell[c]{Model\\revision}} & GGGP & 20.741 & 11.316 & 13.248 & 9.158 \\
			& \best{\textbf{\method}} & \best{\textbf{21.427}} & \best{\textbf{11.966}} & \best{\dunderline{0.7pt}{\textbf{12.356}}} & \best{\dunderline{0.7pt}{\textbf{7.936}}} \\
			\bottomrule
		\end{tabular}
	}
	\label{tab:exp:results:summary}
\end{table}

\subsection{\ref{sec:exp:q1}. Prediction Accuracy}
\label{sec:exp:accuracy}

We split the data into two periods, 1996--2005 for training and 2006--2008 for testing, and
report the forecasting accuracy in~\Cref{tab:exp:results:summary},
in terms of the best RMSE and MAE 
where best models denote those with the smallest test RMSE.

\manual performed significantly worse than other approaches, 
although it is designed with domain knowledge of the biological process and a careful selection of parameter values.
Model calibration approaches, such as \ga, \lhs, and \sa, obtained a much better result than \manual, 
indicating the benefits of tuning model parameters.
However, model calibration methods were outperformed by model revision methods
(with \method obtaining 32\% and 34\% smaller RMSE and MAE than the best model calibration results) 
as they can update only the model parameters, but not the model structure.
In both criteria, \method achieved the best testing performance,
with 7\% and 13\% smaller RMSE and MAE, respectively, than the second best method \gggp.

For data driven models, we used two types of input variables. 
Both variants of \rnn and \arimax (denoted by \textsc{S1} and \textsc{All}) performed worse than model calibration and model revision methods.
While \rnn's best test performance was worse than \method,
\rnn could achieve much smaller training RMSE ({\raise.17ex\hbox{$\scriptstyle\sim$}}6.7) than others as training continued.
However, it suffered from overfitting and its test RMSE increased to {\raise.17ex\hbox{$\scriptstyle\sim$}}44.0.
Note that for both \rnn and \arimax, using additional input variables observed at stations other than S1 
did not help improve the performance.
In fact, for predicting phytoplankton at S1, \arimaxAll performed worse than \arimaxBusan.
As measuring stations are located over a wide area (see \Cref{fig:studysite}),
using measurements from distant stations simply as additional input features was not helpful for predicting at S1.
Also, as is typically the case with ecological data,
the dataset is not large enough (2,435 data points for training) for learning complex processes in a purely data-driven manner.
On the other hand, by using prior knowledge, \method can learn an effective model from a small dataset,
which is also consistent with domain knowledge.

\subsection{\ref{sec:exp:q2}. Ecological Analysis}
\label{sec:exp:ecological_analysis}

From the perspective of ecosystem management, it is critical to understand newly added mechanisms (i.e., extensions).
However, many machine learning applications to environmental research employ uninterpretable, black-box models.
In this section, we strive to understand to what extent the extensions reinforce meaningful information and variables,
and enhance the predictive power of the initial process model.

\textbf{Case Study.} We examined the best models and found that several common variables were added to specific processes. 
First, the addition of temperature dependence was frequently observed, particularly in zooplankton, which is an algal grazer.
\eqref{eq:case:v_tmp} shows an example where a revision is highlighted.
\begin{tcolorbox}[ams align,boxsep=0pt,boxrule=0pt,left=3pt,right=3pt,top=0pt,bottom=3pt]
\begin{split}\label{eq:case:v_tmp}
{\delta}_\mathit{Zoo}	& = \left\{ C_\mathit{DZ} \right\} \highlight{\times ~ (4 V_{tmp} + 253.4)} %
\end{split}
\end{tcolorbox}
\noindent
This shows that the temperature is related to the metabolic rate of zooplankton and its mortality, and
plays a key role on the prediction of algal blooms.

Second, pH was often connected with algal growth process, as in the following example:
{\small
\begin{tcolorbox}[ams align,boxsep=0pt,boxrule=0pt,left=-3pt,right=3pt,top=3pt,bottom=3pt]
\begin{split}\label{eq:case:v_ph}
\frac{d {B_{\mathit{Phy}}}}{d t}\!\!=\!\left\{ B_\mathit{Phy}\!\cdot\!({\mu}_\mathit{Phy}\!-\!{\gamma}_\mathit{Phy})\!-\!B_\mathit{Zoo}\!\cdot\!\varphi \right\} \highlight{+ \frac{V_{alk}}{ V_{ph}\!-\!V_{cd}\!+\!848.4 }}
\end{split}
\end{tcolorbox}}
\noindent
In modeling algal process in rivers, pH has not been often used in knowledge-based modeling
as it is not known to be a limiting factor. %
Although pH is closely associated with aquatic carbon complex 
(which is also an essential component of photosynthesis, along with nitrogen and phosphorus), 
it has been neglected in modeling due to a plethora of carbon availability.
However, it is remarkable to observe the improvement of predictive power when pH is considered as an input variable.
In fact, recent machine learning models have illuminated pH as a crucial factor to predict algal blooms. 
This is one of the major discoveries by \method in the context of river modeling,
which corroborates the importance of pH in environmental research.
Also, note that $ V_{alk} $ is determined by $ CO_3^{2-} $, which is in turn related to pH, and
$ V_{cd} $ is a reasonable selection as it is considered a proxy of pollutant concentration in freshwaters.

\begin{figure}[t!]
	\par\vspace{-1.0em}\par
	\centering
	\includegraphics[width=0.90\linewidth]{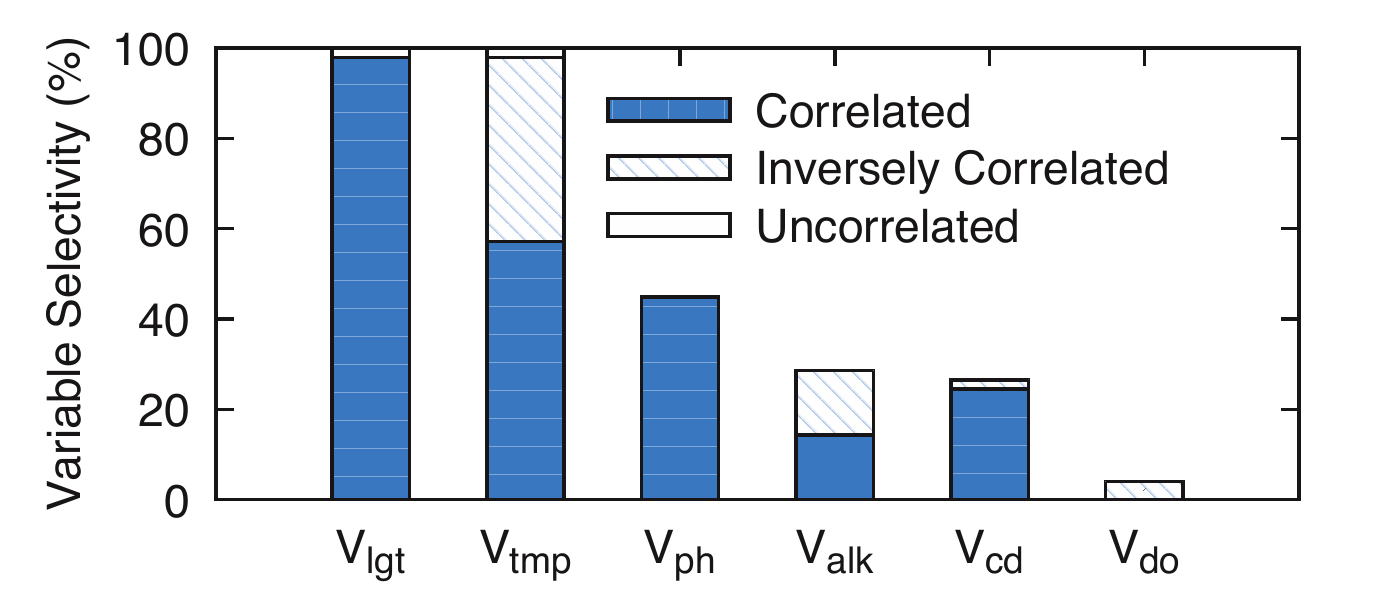}
	\caption{Selectivity of variables among the 50 best models.}
	\label{fig:variable_analysis}
	\vspace{-1em}
\end{figure}

\textbf{Relative Importance Analysis.}
To assess the relative importance of input variables, %
we explored the selectivity (\%) of variables among the 50 best models, and 
the correlation of each variable with phytoplankton ($ B_{Phy} $) growth via variable perturbation (\Cref{fig:variable_analysis}).
Among them, light ($V_{lgt}$) and water temperature ($V_{tmp}$) were selected most often;
this agrees with the fact that they are limiting factors for phytoplankton growth.
Note that while $V_{lgt}$ was consistently correlated with $ B_{Phy} $, $V_{tmp}$ was not.
As water temperature has multiple optimal points for the best growth of $ B_{Phy} $,
$V_{tmp}$ can affect $ B_{Phy} $ either positively or negatively
depending on the dominance of ambient phytoplankton functional group.
The third important factor, $V_{ph}$, is closely related to photosynthesis and oxygen release.
Alkalinity ($V_{alk}$) and electric conductivity ($V_{cd}$) were selected similarly often:
$V_{cd}$'s high correlation is plausible as conductivity can be a proxy of nutrient levels in freshwater.
Dissolved oxygen ($V_{do}$) was negatively correlated with $ B_{Phy} $, 
implying that phytoplankton biomass is lower in higher $V_{do}$ concentrations. 
Overall, while a few exceptions exist, %
most selected ones were ecologically plausible and interpretable.

\begin{figure}[t!]
	\centering
	\makebox[0.4\textwidth][c]{
		\includegraphics[width=0.99\linewidth]{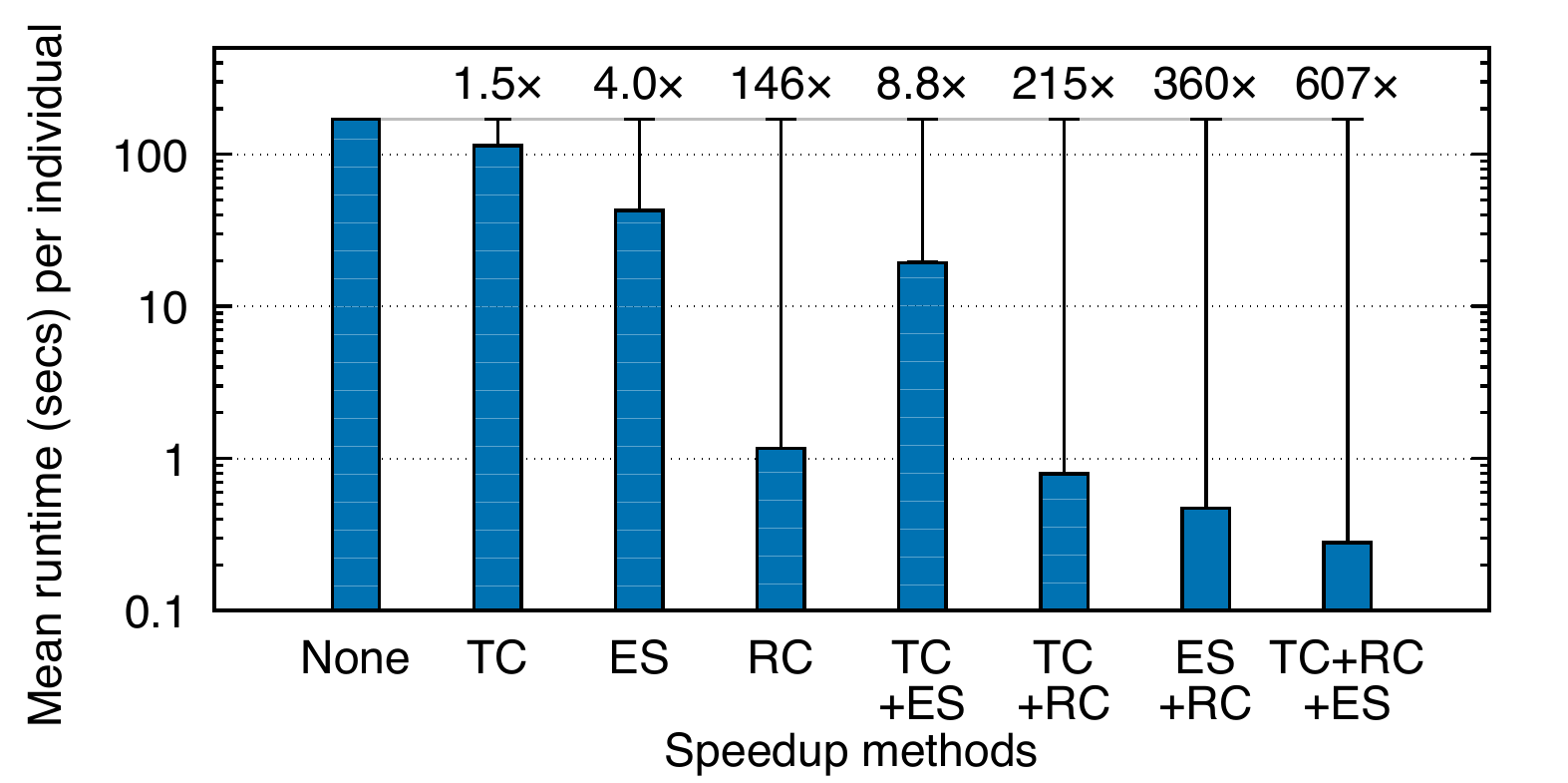}
	}
	\caption{Mean runtime (seconds) per individual by speedup techniques.
		TC: Tree Caching, ES: Evaluation Short-Circuiting, RC: Runtime Compilation. Applying all leads to $ 607\times $ speedup.}
	\label{exp:runtime}
	\vspace{-0.5em}
\end{figure}

\vspace{-1em}
\subsection{\ref{sec:exp:q3}. Analysis of Speedup Techniques}
\label{sec:exp:speedup}

\begin{figure}[t!]
	\centering
	\makebox[0.5\textwidth][c]{
		\includegraphics[width=1.05\linewidth]{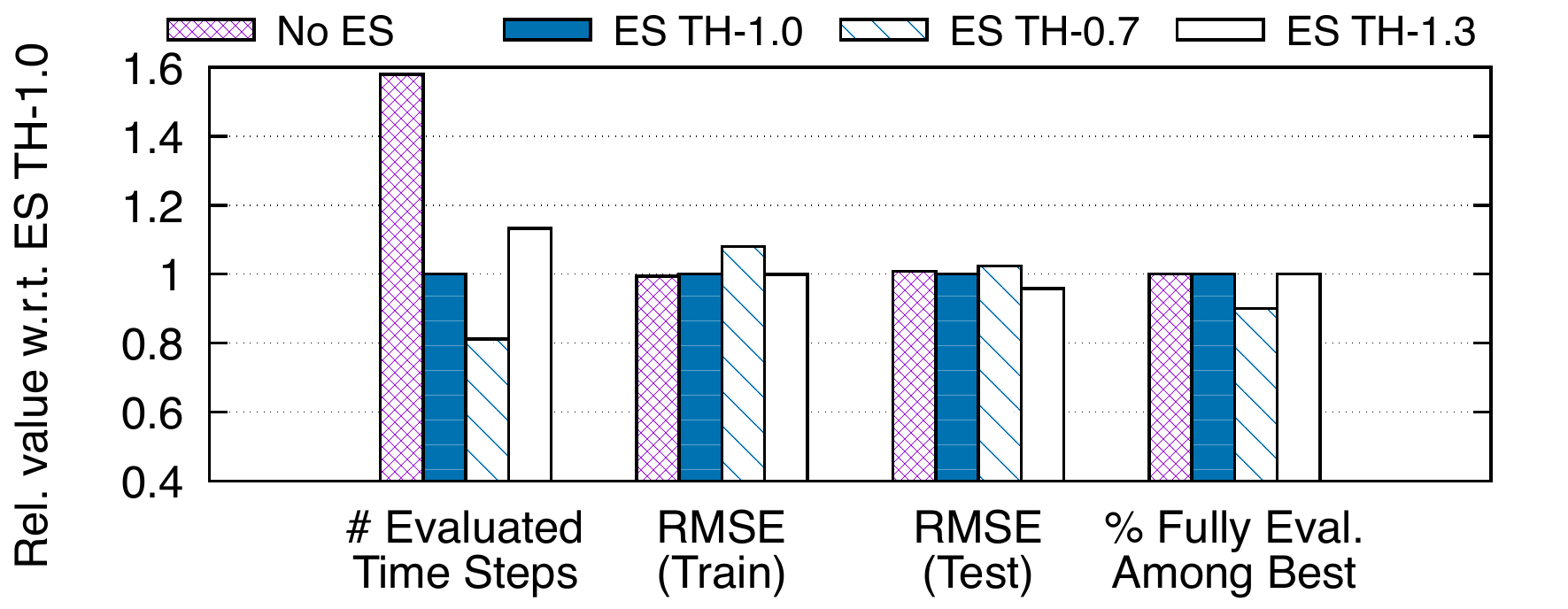}
	}
	\caption{Effect of evaluation short-circuiting (ES) as thresholds (TH) are varied. Relative values w.r.t. ES TH-1.0 are reported.}
	\label{exp:earlytermination}
	\vspace{-1em}
\end{figure}

We apply three orthogonal methods for speedup, i.e., 
tree caching (TC), evaluation short-circuiting (ES), and runtime compilation (RC).
\Cref{exp:runtime} shows the mean runtime (seconds) per individual
obtained with different speedup methods,
which indicates that these techniques effectively reduce computational costs,
achieving 607$ \times $ speedup when all methods are applied together,
compared to when no speedup techniques were used.
We also measured how ES affects the performance when we use different thresholds (0.7, 1.0, and 1.3).
\Cref{exp:earlytermination} shows relative values w.r.t. ES with a threshold of 1.0.
Results indicate that ES generally cut down evaluation cost without sacrifice in accuracy, and
nearly 100\% of the best models were fully evaluated.
While the overall RMSE increased by 5\% as ES got eager (with a threshold of 0.7), the number of evaluated time steps dropped by 19\%.

%% file: 050relatedwork.tex
\textbf{Scientific Discovery from Data.}
Interest in computational methods
for making scientific discoveries from data is growing.
Among data-driven approaches, symbolic regression (SR) using GP~\cite{schmidt2009distilling}
aims to discover scientific laws from data without relying on prior knowledge.
While SR searches for the form of equations and their parameters simultaneously as in genetic model revision (\method),
SR may learn models that are inconsistent with domain knowledge
due to the lack of guidance in the optimization process.

Combining prior knowledge and data science~\cite{DBLP:journals/tkde/KarpatneAFSBGSS17} 
is an emerging paradigm with promising results, which can address this challenge.
One approach along this line is to guide the learning algorithm, e.g., via theory-guided constrained optimization~\cite{DBLP:journals/corr/abs-1710-11431,DBLP:journals/tkde/KarpatneAFSBGSS17}.
\method also falls in this category, using TAG formalism 
for knowledge-guided navigation of the search space.
Model calibration~\cite{DBLP:journals/tkde/KarpatneAFSBGSS17} is another such approach: 
data are used to optimize the parameters of a knowledge-based model.
\method outperforms various model calibration approaches, 
while learning knowledge-consistent processes.

\textbf{Modeling River Water Quality.}
QUAL2E~\cite{brown1987enhanced} is the most well-known model for river ecosystems.
Despite its wide use, QUAL2E's accuracy was limited due to their assumption on steady-state flow.
Neural network (NN) and genetic algorithm (GA) have recently been used for river modeling.
NN-based methods perform data-driven modeling using multilayer perceptron~\cite{singh2009artificial} and 
recurrent neural networks~\cite{kim2018stable}.
By performing model calibration, 
GA-based methods~\cite{DBLP:conf/cec/KimMSLN10,kim2014improvement}
successfully improved the accuracy of a knowledge-based river model.
However, existing evolutionary methods do not have mechanisms to jointly optimize the structure and parameters of a model, 
in particular, guided by domain knowledge.
To fill this gap, we design novel mechanisms to represent prior knowledge and perform knowledge-guided model revision in TAG3P. 
As the first work on modeling river water quality using knowledge-guided model revision,
our work improves upon earlier works that made limited or no use of prior knowledge.

%% file: 060conclusion.tex
Model revision is an effective approach for modeling real-world phenomena 
where domain expertise and data are used simultaneously to model complex dynamical systems.
Our genetic model revision (\method) framework performs model revision guided by prior knowledge.
The case study of river modeling shows its effectiveness.
In future work, we will explore new mechanisms to incorporate domain knowledge 
(new search operators and language biases), and 
apply \method to other domains, such as financial forecasting.

\textbf{Reproducibility}: Code and data are available at \url{https://www.cs.cmu.edu/~namyongp/gmr}.

\textbf{Extensibility}: How readily can the ideas behind this work be extended to other domains? The degree of effort required depends on the similarity to the present problem. At one extreme, the underlying ideas are applicable to most model identification problems where expert knowledge is available but incomplete. TAG grammars are particularly suited to this, because they readily match the common situation where experts have a simple model that they believe to be generally correct, but potentially to require modification because potentially important processes have been omitted for simplicity. The adjunction operation of TAG3P is particularly suited to recording the way experts think about such problems. Similarly, it will frequently be the case that experts can define feasibility bounds on model parameters. In real world problems, it is frequently the case that evolving models need to be repeatedly evaluated to estimate their fitness. In such cases, the speedup techniques we have described here may well be applicable. In all these circumstances, portions of the code we have made available may be useful for implementation.

At the other extreme, the system can be directly applied to other river systems under the assumption of conservative, non-branching flow, provided that the corresponding data is available. The system also relies specifically on the G++ compiler suite to provide run-time compilation. Similar capabilities are available in other C++ compilers, or run-time compilation can be sacrificed at a substantial cost in speed.

Where the data differs in the water properties collected, revisions in the search limiting grammar will be required, and prior knowledge of the feasible values of any new parameters will need to be specified.

Extensions to river systems with substantial water loss through evaporation or leakage, or to braided streams or delta, require significant re-programming of the flow model, and additional data inputs for those components. 

Moving beyond rivers, much of the structure of the system is determined by the need to model variables that change as a fluid flows through a network. It is not hard to find other problems that fit that bill: flow of blood through an organism; flow of feedstocks through a chemical plant; municipal water or sewage systems. In all these cases, the existing code would form a useful starting point.

%% file: 070appendix.tex
\section{Further Details of River Modeling}\label{sec:rivermodeling}

\begin{figure}[!t]
	\centering
	\includegraphics[width=0.6\linewidth]{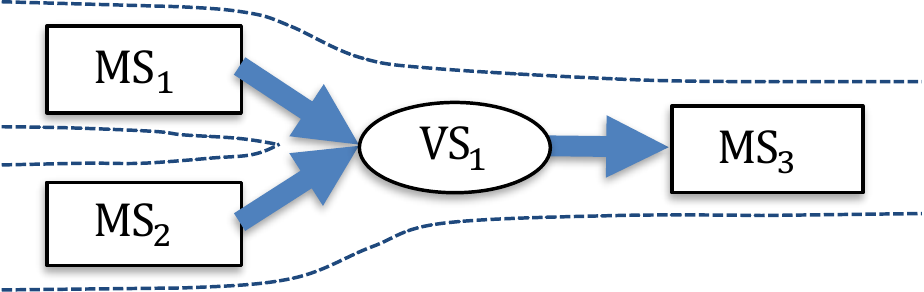}
	\caption{An example river system with measuring stations (MS) and a virtual station (VS).
		VS is placed at a confluence.}
	\label{fig:stations}
\end{figure}

\textbf{Modeling A River System.} 
To monitor the ecological status of a river,
sample measurement is performed, often on a regular basis, 
at multiple measuring stations located in geographically important places.
Based on this data collection scheme,
we model a river system as a directed acyclic graph as shown in~\Cref{fig:stations},
where a node corresponds to a measuring station and an edge denotes a segment of a river
between the two adjacent stations.
To model the confluence, we add a virtual station where two or more water bodies meet together as depicted in~\Cref{fig:stations} (VS\textsubscript{1}).
We model our study site shown in \Cref{fig:studysite} 
by adding six stations (S1--S6) at the main channel of the river,
three stations (T1--T3) at the major tributaries of the river, and
three virtual stations at the confluence of a tributary with the main stream (S6 $ \cdot $ T3, S4 $ \cdot $ T2, and S3 $ \cdot $ T1).

\textbf{Hydrological Process.}
Modeling a river system involves modeling two contemporaneous processes, the \textit{biological} and \textit{hydrological} processes. 
The biological process shown in~\eqref{eq:biologicalP:raw} and \eqref{eq:biologicalZ:raw} models 
the evolution of phytoplankton and zooplankton and their interaction.
The hydrological process models the flow of water bodies, and 
provides information of flow at specific time to the biological process.
We use the hydrological process first introduced in~\cite{DBLP:conf/cec/KimMSLN10},
which is based on a flow mass balance between stations.
Specifically, given that water flows from station A to station B,
flow into B consists of three components, (i) inflow from upper station A, 
(ii) water retained at B (e.g., due to water trapped in side pools or non-laminar flow), and (iii) runoff into B from precipitation:
{
\setlength{\abovedisplayskip}{3pt}
\setlength{\belowdisplayskip}{\abovedisplayskip}
\setlength{\abovedisplayshortskip}{0pt}
\setlength{\belowdisplayshortskip}{3pt}
\begin{align}\label{eq:flowmodel}
F_{B,t+\Delta} = r_B \cdot F_{B,t} + (1 - r_A) \cdot F_{A,t} + R_{B,t+\Delta}
\end{align}}where $ F_{S,t} $ denotes the flow at station $ S $ at time $ t $,
$ r_S $ is the ratio of the water retained at station $ S $,
$ \Delta $ is the time taken for water bodies from A to arrive at B, and
$ R_{S,t} $ denotes the amount of inflow into station B at time $ t $ that arises from rain fall.
Our hydrological process also models how water bodies are merged at a confluence (e.g., VS\textsubscript{1} in~\Cref{fig:stations}),
in which case the attributes (e.g., nutrient level) of each water body are updated based on~\eqref{eq:flowmodel} 
as they are moving to the confluence, and the water bodies are aggregated as a flow-weighted average.

Thus, the hydrological process determines the attributes of a water body at a specific location and time
by modeling how water bodies are merged, how the rainfall is absorbed into the river, etc.
As the attributes of a water body are used by the biological process,
the hydrological process affects the biological process.
As we focus on modeling the biological process, we use a static hydrological process in this work. %

\textbf{Biological Process.}
We evolve a population of individuals (\Cref{fig:framework}),
where each one is a revised biological process.
Each individual receives values for the temporal variables (\Cref{tab:process_variables})
from the water body that the hydrological process provides, 
and updates the phytoplankton ($ B_{Phy} $) and zooplankton ($ B_{Zoo} $) according to its process equation.
\Cref{tab:process_variables,tab:process_constants} list the variables used in our biological process.

\vspace{-0.5em}
\section{Experimental Settings}
\label{sec:exp:settings}

\textbf{Environment.}
We used an Ubuntu server with 80 Intel Xeon E7-4850 processors at 2.00GHz and 252GB RAM.

\textbf{\method, \gggp, and \ga.} 
We implemented \method, \gggp, and \ga frameworks in C++.
For \method, we used the following configurations:
number of generations (100), population size (200), number of runs (60), minimization objective (RMSE),
elite size (2), number of local search steps (5), selection mechanism (tournament selection), tournament size (5),
minimum chromosome size (2), maximum chromosome size (50).
We applied the following genetic operators (the number in the parentheses denotes the operator probability):
crossover (0.3), subtree mutation (0.3), Gaussian mutation (0.3), and replication (0.1).
For \gggp and \ga, we used the same configurations used for \method.
Since local search incurs additional fitness evaluation in \method, 
\gggp and \ga used a population of 1200 individuals
to use the same number of fitness evaluation for both methods.

\textbf{\rnn.} 
We used a two-layer LSTM implemented in PyTorch, whose hidden size was equal to the number of input features.
The output was transformed into an estimated phytoplankton value via dense neural networks with two layers.
The input features were standardized.
We used Adam optimizer with $ \alpha=0.01, \beta_1 = 0.9, \beta_2 = 0.999 $, and weight decay of 0.0005.
The model was trained for up to 1000 epochs with MSE loss.

\textbf{\arimax.} We used the \textit{pmdarima} library and its Auto-ARIMA functionality with the default parameter settings.

\textbf{Others.} For other model calibration methods, we used the SPOTPY framework~\cite{houska2015spotting} using RMSE as the objective function. 
We set method parameters to their default values.